\documentclass[conference]{IEEEtran}
\usepackage{cite}
\usepackage{amsmath,amssymb,amsfonts}
\usepackage{graphicx}
\usepackage{textcomp}
\usepackage{bmpsize}
\usepackage{xcolor}
\usepackage{lipsum}

\usepackage{graphicx}
\usepackage{booktabs}
\usepackage{subcaption}
\usepackage{amsmath}
\usepackage{multirow}

\usepackage{booktabs}
\usepackage{multirow}
\usepackage{xcolor}
\usepackage{makecell}
 \usepackage{hyperref}
\usepackage{soul}
\usepackage{xcolor}
\usepackage{bm}
\usepackage{soul}
\usepackage{adjustbox}
\usepackage{tcolorbox}
\usepackage{colortbl}
\usepackage[inline]{enumitem}
\usepackage{pgfplots}
\pgfplotsset{compat=1.18}

\usepackage{algorithm}
\usepackage{algorithmic}

\usepackage{amsmath}

\newtheorem{theorem}{Theorem}  
\newtheorem{corollary}{Corollary}  

\def\BibTeX{{\rm B\kern-.05em{\sc i\kern-.025em b}\kern-.08em
    T\kern-.1667em\lower.7ex\hbox{E}\kern-.125emX}}
\begin{document}


\title{\texttt{ADAPT:} A Pseudo-labeling Approach to Combat Concept Drift in Malware Detection}


\author{
\IEEEauthorblockN{1\textsuperscript{st} Md Tanvirul Alam}
\IEEEauthorblockA{\textit{Rochester Institute of Technology} \\
Rochester, NY, USA \\
ma8235@rit.edu}
\and
\IEEEauthorblockN{2\textsuperscript{nd} Aritran Piplai}
\IEEEauthorblockA{\textit{University of Texas El Paso} \\
El Paso, TX, USA \\
apiplai@utep.edu}
\and
\IEEEauthorblockN{3\textsuperscript{rd} Nidhi Rastogi}
\IEEEauthorblockA{\textit{Rochester Institute of Technology} \\
Rochester, NY, USA \\
nxrvse@rit.edu}
}

\date{July 11, 2025}  

\maketitle
\begin{abstract}
Machine learning models are commonly used for malware classification; however, they suffer from performance degradation over time due to concept drift. Adapting these models to changing data distributions requires frequent updates, which rely on costly ground truth annotations. While active learning can reduce the annotation burden, leveraging unlabeled data through semi-supervised learning remains a relatively underexplored approach in the context of malware detection. In this research, we introduce \texttt{ADAPT}, a novel pseudo-labeling semi-supervised algorithm for addressing concept drift. Our model-agnostic method can be applied to various machine learning models, including neural networks and tree-based algorithms. We conduct extensive experiments on five diverse malware detection datasets spanning Android, Windows, and PDF domains. The results demonstrate that our method consistently outperforms baseline models and competitive benchmarks. This work paves the way for more effective adaptation of machine learning models to concept drift in malware detection.

\end{abstract}


\section{Introduction}


The ever-evolving nature of malware poses ongoing challenges to cybersecurity. While machine learning offers substantial improvements over traditional signature- and rule-based detection~\cite{arp2014drebin,2018arXiv180404637A,DBLP:journals/tissec/OnwuzurikeMACRS19,rabadi2020advanced,issakhani2022pdf}, its effectiveness depends on the assumption that data distributions remain stable~\cite{bishop2006pattern}. In reality, continuous changes in malware and underlying platforms—such as new features, APIs, and programming practices~\cite{kan2021investigating}—frequently violate this assumption, causing \textit{concept drift} and leading to significant model degradation unless adaptation measures are taken~\cite{lu2018learning,pendlebury2019tesseract}.


Dealing with concept drift in malware detection is a multifaceted challenge. While some approaches focus on building more robust feature spaces to resist drift~\cite{DBLP:journals/tissec/OnwuzurikeMACRS19,zhang2020enhancing}, true immunity remains elusive given malware’s dynamic landscape~\cite{barbero2022transcending}. Periodic model retraining offers another solution~\cite{narayanan2017context,pendlebury2019tesseract}, but requires timely drift detection and access to newly labeled samples~\cite{jordaney2017transcend,barbero2022transcending,yang2021cade}. Active learning helps reduce annotation effort~\cite{nissim2014novel,chen2023continuous}, yet obtaining high-quality malware labels is costly due to the sheer volume of emerging malware variants~\cite{miller2016reviewer,braun2024understanding}.


Self-training offers a way to adapt to concept drift without requiring new annotations beyond the initial training data~\cite{amini2022self}. Pseudo-labeling is a learning paradigm where the model generates labels for unlabeled data, treating its predictions as ground truth to train itself~\cite{lee2013pseudo, shi2018transductive} further. A notable application of this approach for malware detection is DroidEvolver~\cite{xu2019droidevolver}, which employs pseudo-labeling for Android malware detection by using an ensemble of online learning algorithms to eliminate labeling costs after the initial training phase. However, a later study~\cite{kan2021investigating} revealed that when experimental biases, such as unrealistic benign-to-malware ratios in the dataset, are addressed, DroidEvolver's performance deteriorates significantly due to confirmation bias~\cite{arazo2020pseudo} and a catastrophic self-poisoning effect, leading to a sudden and severe performance drop. Although the work in~\cite{kan2021investigating} proposed an enhanced version of DroidEvolver to mitigate these issues, performance still deteriorates rapidly after a short period when solely relying on pseudo-labels. Thus, designing a practical pseudo-labeling-based approach for malware detection under concept drift remains a significant challenge.


In this work, we propose \texttt{ADAPT}: \textbf{A}daptive \textbf{D}rift-aware \textbf{A}lgorithm using \textbf{P}seudo-labeling for malware detec\textbf{T}ion, a semi-supervised approach tailored for concept drift in malware detection. Unlike general-purpose methods, \texttt{ADAPT} accounts for the asymmetric nature of drift, which more frequently impacts malware than benign samples~\cite{barbero2022transcending,chow2023drift}, by introducing a drift-aware pseudo-labeling strategy that selectively propagates labels based on class-specific drift behavior. Data augmentation, though standard in image classification~\cite{cubuk2020randaugment}, remains uncommon in malware detection, but has shown promise in noisy label settings~\cite{wu2023grim}. We extend these augmentation strategies to the concept drift scenario, periodically retraining the model to adapt to changing distributions. To further enhance robustness, we incorporate mixup regularization~\cite{mixup, thulasidasan2019mixup}, which improves confidence calibration and helps mitigate the self-poisoning effect commonly seen in pseudo-labeling. This ensures that the model’s predicted probabilities better reflect the actual likelihood of correctness, making pseudo-labeling more effective in practice.

We evaluate \texttt{ADAPT} on five diverse malware detection datasets—two Android, two Windows, and one PDF—and show that it consistently outperforms competitive semi-supervised learning methods, achieving state-of-the-art results across all tasks. We also demonstrate its effectiveness in an active learning setting, where \texttt{ADAPT} surpasses prior state-of-the-art performance on Android malware detection under the same annotation budget.

Our key contributions are:

\begin{enumerate}
\item We introduce \texttt{ADAPT}\footnote{\url{https://github.com/aiforsec/ADAPT}}, a semi-supervised learning algorithm tailored for concept drift in malware detection. It mitigates catastrophic self-poisoning through drift-aware pseudo-labeling.

\item Our method is model-agnostic and shows improvements across Random Forest, XGBoost, and neural networks. We validate its effectiveness on five real-world malware datasets exhibiting varying degrees of concept drift.

\item We extend \texttt{ADAPT} to an active learning setting, achieving state-of-the-art results on Android malware detection. We also apply it to the multiclass problem of malware family classification and demonstrate its effectiveness under concept drift.
\end{enumerate}
\section{Background and Related Work}
\label{related}

\subsection{Malware Detection using Machine Learning}
Machine learning has been widely applied to malware detection~\cite{chen2020training, arp2014drebin, 2018arXiv180404637A, lindorfer2015marvin, DBLP:journals/tissec/OnwuzurikeMACRS19, zhang2020enhancing}, both in academic research and industrial settings. These models are trained using features extracted through static analysis, dynamic analysis, a combination of both, or memory-based analysis~\cite{sihwail2018survey}. Static features are obtained by analyzing the binary or source code of a file without executing it, making them a popular choice for malware detection~\cite{arp2014drebin, DBLP:journals/tissec/OnwuzurikeMACRS19, 2018arXiv180404637A}. While machine learning detectors built upon static features often achieve high performance when the training and test datasets follow the same distribution, their effectiveness deteriorates in real-world scenarios due to distributional changes caused by code obfuscation, concept drift, and adversarial examples~\cite{gao2024comprehensive}. In particular, concept drift, also known as dataset shift,  is the primary focus of this study.

\subsection{Concept Drift}

Dataset shift refers to the phenomenon in machine learning where the joint distribution of inputs and outputs differs between the training and test stages~\cite{quinonero2022dataset}. Dataset shifts can be broadly categorized into three types: covariate shift, label shift, and concept shift. Covariate shift occurs when the distribution of the features, \( p(x) \), changes. Label shift, also known as prior probability shift, refers to a change in the distribution of the labels, \( p(y) \). Concept shift refers to a change in the conditional probability distribution, \( p(y | x) \), meaning the relationship between the features and the target labels changes, effectively altering the class definitions. The individual effects of these shifts can be challenging to disentangle from a finite set of samples~\cite{quinonero2022dataset,barbero2022transcending}. Consequently, in machine learning for security, it is common practice to refer to all these shifts under the broader term \textit{concept drift}~\cite{jordaney2017transcend,kan2021investigating,barbero2022transcending}, which is the terminology we adopt in this work.

Concept drift in malware detection arises from both benign and malicious changes. Updates to APIs and the introduction of new functionalities in benign applications contribute to changes in the input feature space, often resulting in what is referred to as \emph{virtual drift}~\cite{shyaa2024evolving}. In contrast, the primary driver of concept drift is the continually evolving behavior of malware~\cite{chen2023overkill,zhang2020enhancing}, which leads to \emph{real concept drift}, a shift in the relationship between features and labels~\cite{shyaa2024evolving}. Adversaries regularly adapt through obfuscation, packing, and novel attack strategies to evade detection~\cite{aghakhani2020malware,anderson2018learning,tam2017evolution}. These dynamics alter the distribution of malicious behavior between training and testing data~\cite{barbero2022transcending}, degrading classifier performance and reducing detection rates over time~\cite{pendlebury2019tesseract,jordaney2017transcend}.

\subsection{Concept Drift Adaptation}

While developing feature spaces that are inherently more resilient to concept drift is an effective approach~\cite{DBLP:journals/tissec/OnwuzurikeMACRS19, zhang2020enhancing}, the dynamic nature of malware makes it challenging to design such feature spaces that remain effective over long periods~\cite{barbero2022transcending}. As a result, machine learning models trained on these features need to be periodically retrained to stay functional.

Approaches such as Transcend~\cite{jordaney2017transcend} and Transcendent~\cite{barbero2022transcending} address concept drift by employing classification with rejection, using conformal evaluators to flag uncertain samples for expert review. However, reliance on manual labeling inherently limits the frequency of model retraining due to cost and resource constraints~\cite{miller2016reviewer,xu2019droidevolver}. Active learning has been proposed to alleviate the annotation burden by selectively labeling only the most uncertain samples~\cite{yang2021bodmas,chen2023continuous}, with recent studies demonstrating up to an eightfold reduction in labeling costs~\cite{chen2023continuous}. 

Despite these advances, practical challenges such as label delays can significantly increase user exposure to undetected malware~\cite{botacin2025towards}. As a result, there is a clear need for interim solutions that can adapt to drift in the absence of timely ground-truth labels. Pseudo-labeling provides one such solution, enabling models to update themselves using confident predictions in place of manual annotations~\cite{botacin2025towards,xu2019droidevolver,kan2021investigating}.

\subsection{Pseudo-Labeling for Drift Adaptation}
Pseudo-labeling is a technique that leverages model predictions to assign labels to unlabeled data, assuming that predictions with high confidence are likely to be accurate. These pseudo-labels are then incorporated into the training process, enabling the model to learn from labeled and unlabeled samples. Pseudo-labeling is widely used in semi-supervised learning setups~\cite{sohn2020fixmatch, berthelot2019mixmatch}. However, in the context of malware detection, its application typically does not account for adapting to concept drift~\cite{wu2023grim, mahdavifar2020dynamic, santos2011semi, noorbehbahani2020ransomware}.

DroidEvolver~\cite{xu2019droidevolver} applies pseudo-labeling for Android malware detection under concept drift using an ensemble of linear models and online learning to avoid manual labeling. However, it was later shown to underperform when accounting for dataset biases~\cite{kan2021investigating}, prompting several improvements. More recent work includes MORPH~\cite{alam2024morph}, which uses neural networks but suffers from high false positive rates, and Insomnia~\cite{andresini2021insomnia}, which applies co-training for drift adaptation in network intrusion detection.

Despite its potential, pseudo-labeling under concept drift poses challenges: as distributions shift, model predictions become less reliable, leading to incorrect pseudo-labels and self-reinforcing errors—known as self-poisoning~\cite{arazo2020pseudo, kan2021investigating}. In this work, we propose a pseudo-labeling approach specifically designed to mitigate these effects.
\section{Research Motivation}
\label{sec-motivation}

\begin{figure}[t]
  \centering
  \begin{subfigure}[b]{0.24\textwidth}
    \includegraphics[width=\textwidth]{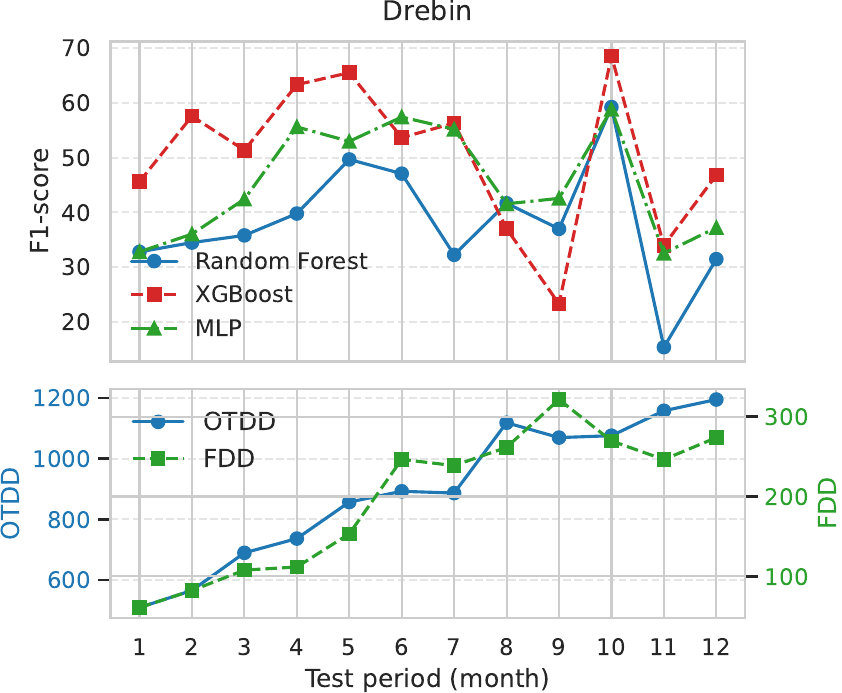}
  \end{subfigure}
  \hfill
  \begin{subfigure}[b]{0.24\textwidth}
    \includegraphics[width=\textwidth]{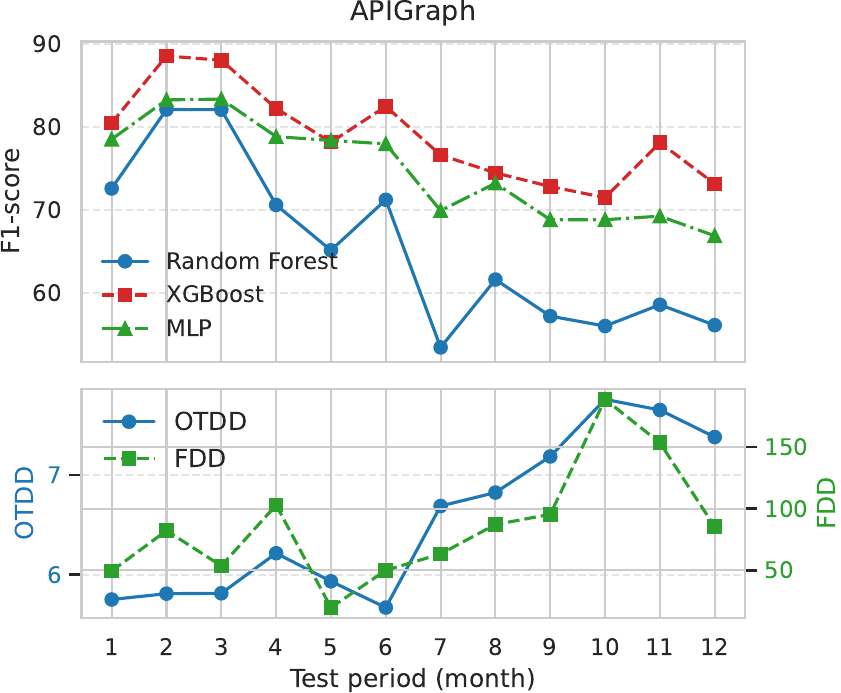}
  \end{subfigure}
\caption{Covariate shift \( p(x) \) and concept shift \( p(y|x) \) measured by Optimal Transport Dataset Distance (OTDD) and Fréchet Dataset Distance (FDD) across the first year of testing on two Android datasets (Drebin and APIGraph). The corresponding monthly F1-scores from three models (Random Forest, XGBoost, and MLP) are shown to illustrate the relationship between distributional shifts and malware detection performance.}
  \label{fig:concept-drift}
\end{figure}

\paragraph{\textbf{Nature of Concept Drift in Malware Detection}}
Machine learning models used for malware detection experience degraded performance over time due to concept drift. Although precise measurement of dataset shifts from finite samples is challenging, they can be approximated using metrics such as the Optimal Transport Dataset Distance (OTDD)~\cite{alvarez2020geometric} for covariate shift and the Fréchet Dataset Distance (FDD)~\cite{heusel2017gans} for concept shift~\cite{gardner2024benchmarking}.

OTDD~\cite{alvarez2020geometric} quantifies covariate shift by calculating the minimal cost to transform one data distribution into another using optimal transport theory. Specifically, we approximate this distance with a Gaussian approximation computed directly in the feature space between the training data and each subsequent month's testing data. In contrast, Fréchet Dataset Distance (FDD)~\cite{heusel2017gans}, initially introduced in the context of generative modeling, captures concept shift by evaluating the similarity of intermediate feature representations extracted from a trained reference classifier. Here, we employ a Multilayer Perceptron (MLP) model trained with optimal hyperparameters (as detailed in Section~\ref{sec:hyperparameters}) and compare penultimate layer features from each test month's samples against those of the training set. 

Applying these metrics to two prominent Android malware datasets, Drebin~\cite{arp2014drebin} and APIGraph~\cite{zhang2020enhancing} (Figure~\ref{fig:concept-drift}), we observe substantial and progressively increasing covariate and concept shifts. These distributional shifts are notably more pronounced in Drebin, which utilizes a larger, less semantically compact feature space compared to APIGraph's more concise API-based representation. Although both datasets experience degradation in malware detection accuracy over time, this deterioration more consistently correlates with increased OTDD and FDD distances in APIGraph, highlighting complex interactions between dataset shifts and classifier performance. These insights underscore the critical importance of continuously adapting models to mitigate the negative impacts of concept drift.

\begin{figure}[t]
\centering
\includegraphics[width=0.48\textwidth]{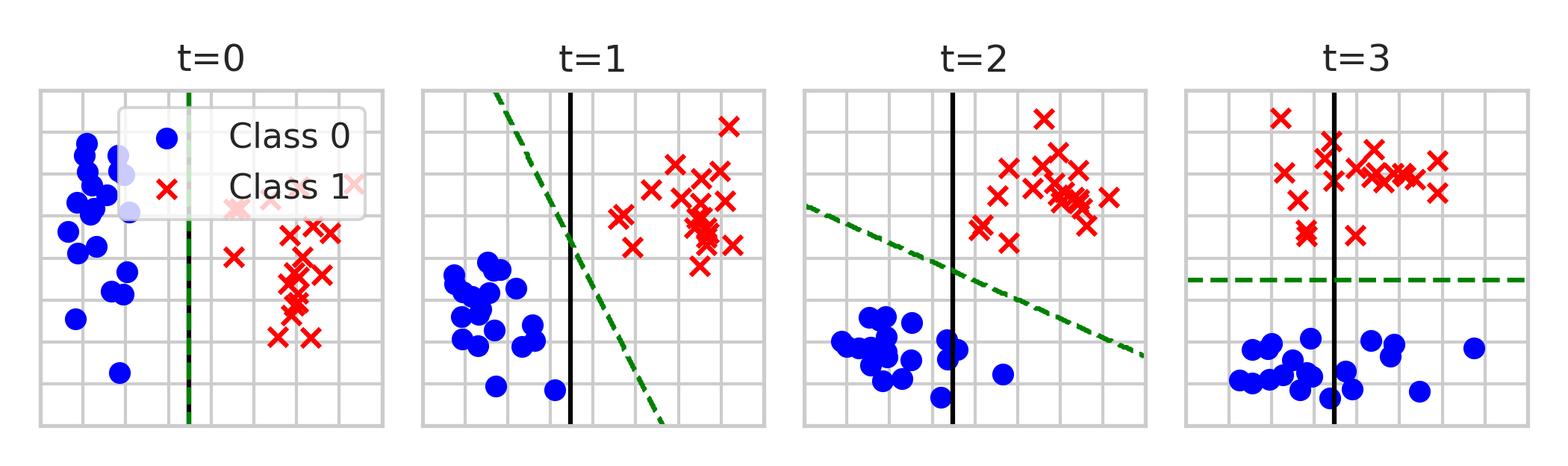}
\caption{Adaptation to distribution shift using pseudo-labels. Solid line: original classifier; dotted lines: adapted classifier trained with pseudo-labels.}
\label{fig:gradual-drift}
\end{figure}

\paragraph{\textbf{Challenges of Label Delays in Practice}}
Effective adaptation to concept drift fundamentally relies on the availability of timely ground-truth labels for evaluating model predictions. However, in practice, these labels often arrive after significant delays due to resource-intensive processes. Sandbox analyses, essential for automated malware evaluation, face scalability constraints relative to the volume of daily submissions, resulting in prolonged analysis queues~\cite{finder2022time}. Human analysts are even scarcer, causing manual labeling to become a severe bottleneck with substantial waiting periods~\cite{beaugnon2018end}. Consequently, these delays open extended vulnerability windows, significantly increasing user exposure to emerging threats and degrading detection performance~\cite{botacin2025towards}.

\paragraph{\textbf{Pseudo-labeling for Effective Drift Adaptation}}
To mitigate these practical challenges associated with label delays, pseudo-labeling emerges as a promising strategy. 
Pseudo-labeling is a self-training technique in which model-generated predictions serve as provisional labels for unlabeled data, enabling continuous model adaptation without immediate reliance on manual annotations~\cite{kumar2020understanding}. This process is illustrated in Figure~\ref{fig:gradual-drift} using a 2D toy dataset with a logistic regression classifier. In this example, both \( p(x) \) and \( p(y|x) \) gradually change from \( t=0 \) to \( t=3 \), with the dataset undergoing continuous rotation. As a result, the classifier’s initial decision boundary (solid line) misclassifies most samples by \( t=3 \). 

However, by using the model’s own predictions as pseudo-labels and retraining on these pseudo-labeled samples, the model can adapt to the changing data distribution. The updated classifier (dashed line) in Figure~\ref{fig:gradual-drift} successfully separates the dataset at the intermediate stages and in the final step. While this example demonstrates the effectiveness of self-training in a simple, idealized scenario, real-world malware datasets are significantly more complex. A naive application of pseudo-labeling in these cases can lead to performance degradation~\cite{kan2021investigating}.

Nevertheless, malware datasets typically undergo gradual distributional shifts, as illustrated in Figure~\ref{fig:concept-drift}, where both covariate and concept shifts increase over time. Based on this observation, we hypothesize that an effective self-training approach can continuously adapt the model to account for these evolving shifts. Notably, prior work by Kumar et al.~\cite{kumar2020understanding} establishes formal error bounds of self-training under the assumption of gradual distributional shifts. Our method builds on this foundation by extending their framework, and we provide a rigorous theoretical analysis in Appendix~\ref{sec:theory}.




\section{Proposed Method}
\label{sec:met}

We introduce \texttt{ADAPT}, a pseudo-labeling-based self-training algorithm for handling concept drift in malware detection. \texttt{ADAPT} is compatible with various learning algorithms, including tree-based methods and neural networks. We evaluate three baseline models in our experiments: Random Forest (RF), XGBoost, and Multilayer Perceptron (MLP). These models were selected because different algorithms may perform better on different datasets, and simpler models like Random Forest can be preferable due to their efficiency in specific applications~\cite{dambra2023decoding,yang2021bodmas,vsrndic2016hidost,jordaney2017transcend,alam2024revisiting}.

We assume that we initially have access to an annotated, labeled dataset. Specifically, we are given a set of $N$ annotated samples with features $\{x_1, x_2, \ldots, x_N\}$ and their corresponding binary labels $\{y_1, y_2, \ldots, y_N\}$, where 0 represents a benign sample and 1 represents malware. We first train a supervised machine learning model, denoted as $M_0$, using this labeled data. We later extend our analysis to the multiclass setting, which is discussed in Section~\ref{sec:multiclass}.

After the initial training, the model is periodically updated each subsequent month using unlabeled test data. This experimental setting aligns with prior work on concept drift adaptation, such as the approach by Chen et al.~\cite{chen2023continuous} for continuous learning. We employ our proposed algorithm during each monthly retraining period to leverage the unlabeled data, improving the model’s ability to adapt to concept drift. 

\begin{algorithm}[t]
\caption{\texttt{ADAPT}: Adaptive Drift-aware Algorithm using Pseudo-labeling for malware detection}
\label{alg:adapt}
\begin{algorithmic}[1]
    \STATE \textbf{Input:} Labeled dataset $\mathcal{D}_l$, unlabeled dataset $\mathcal{D}_u$
    \STATE \textbf{Output:} Updated model $M$

    \STATE Train initial model $M_0$ on the labeled dataset $\mathcal{D}_l$
    \FOR{each test month from $i = 1$ to $T$}
        \STATE \textbf{Step 1: Pseudo-Label Selection}
        \STATE Select pseudo-labeled samples: $\mathcal{D}_p = \{(x_i, \hat{y}_i)\}$ from the unlabeled dataset $\mathcal{D}_u$ using the model $M_{i-1}$ [see \S\ref{sec:pseudo-label}]. 
        \STATE Merge the labeled and pseudo-labeled data: $\mathcal{D}_m = \mathcal{D}_l \cup \mathcal{D}_p$

        \STATE \textbf{Step 2: Data Augmentation}
        \STATE Apply data augmentation to $\mathcal{D}_m$ to generate the augmented dataset $\mathcal{D}_{\text{aug}}$ [\S\ref{sec:augment}].

        \STATE \textbf{Step 3: Mixup Regularization}
        \STATE Apply mixup regularization on $\mathcal{D}_m$ to produce the mixup-augmented dataset $\mathcal{D}_{\text{mixup}}$ [\S\ref{sec:mixup}].

        \STATE \textbf{Step 4: Model Retraining}
        \STATE Combine the merged, augmented, and mixup datasets: $\mathcal{D}_{\text{combined}} = \mathcal{D}_m \cup \mathcal{D}_{\text{aug}} \cup \mathcal{D}_{\text{mixup}}$
        \STATE Retrain the model $M_i$ for the current month on the combined dataset $\mathcal{D}_{\text{combined}}$
    \ENDFOR
\end{algorithmic}
\end{algorithm}

\texttt{ADAPT} comprises four key steps during the semi-supervised update phase:

\begin{enumerate}

    \item \textbf{Adaptive Drift-Aware Pseudo-Labeling:} In this step, we select model predictions that are eligible for inclusion in the self-training process. We propose a novel adaptive pseudo-label selection method to improve the label quality of the selected samples. Details of the pseudo-labeling method are provided in Section~\ref{sec:pseudo-label}.
    
    \item \textbf{Data Augmentation with Label Consistency:} To increase the diversity of the training data, we apply a data augmentation phase. Augmentation has proven to be an effective strategy in self-training and semi-supervised learning, especially in scenarios like malware classification where labels can be noisy. The specifics of our augmentation technique are discussed in Section~\ref{sec:augment}.
    
    \item \textbf{Confidence Calibration with Mixup:} High-quality pseudo-labels are crucial for successful self-training. Poor pseudo-labels can lead to issues such as self-poisoning, where the model reinforces its incorrect predictions, and confirmation bias. Since we use confidence thresholding to filter out low-confidence predictions, it is essential to calibrate the model's confidence levels—i.e., to ensure that predictions with high confidence are indeed more likely to be correct. To achieve this, we integrate the mixup~\cite{mixup} regularization technique, which has been shown to improve model calibration in prior work~\cite{thulasidasan2019mixup}. Further details on the mixup method are provided in Section~\ref{sec:mixup}.
    
    \item \textbf{Model Retraining:} In the final step, we combine the original labeled data, the pseudo-labeled data, and the datasets generated through augmentation and mixup regularization. The model is then updated on this combined dataset. For traditional models like random forests and XGBoost, we retrain the model from scratch, while for neural networks, we fine-tune the model, following the guidelines in~\cite{chen2023continuous}.
    
\end{enumerate}

Algorithm~\ref{alg:adapt} provides an overview of the \texttt{ADAPT} algorithm.

\subsection{Adaptive Drift-Aware Pseudo-Labeling}
\label{sec:pseudo-label}

In traditional pseudo-label selection, a fixed threshold $\tau$ is used to filter out low-confidence predictions~\cite{lee2013pseudo}. Given a model $M$ and its predicted probability for a sample $\mathbf{x}_j$, the pseudo-label $\tilde{y}_j$ is defined as:

\[
\tilde{y}_j =
\begin{cases}
\arg\max_{k} \hat{y}_j^{(k)}, & \text{if } \max_{k} \hat{y}_j^{(k)} \geq \tau, \\
\text{unlabeled}, & \text{otherwise}.
\end{cases}
\]

Here, $\hat{y}_j^{(k)}$ denotes the predicted probability of the $k$-th class for the unlabeled sample $\mathbf{x}_j$. This formulation assumes that the model outputs probabilities for two classes, benign and malware, which sum to 1. If the model instead outputs only the probability $p_1$ for the positive (malware) class—such as in the case of a neural network with a single output neuron and sigmoid activation—then the probability for the benign class is $p_0 = 1 - p_1$. In either case, if the predicted probability for the assigned class exceeds the predefined threshold $\tau$, the sample is pseudo-labeled; otherwise, it remains unlabeled.

\begin{figure}[t]
    \centering

    \begin{subfigure}[t]{0.99\linewidth}
        \centering
        \includegraphics[width=\linewidth]{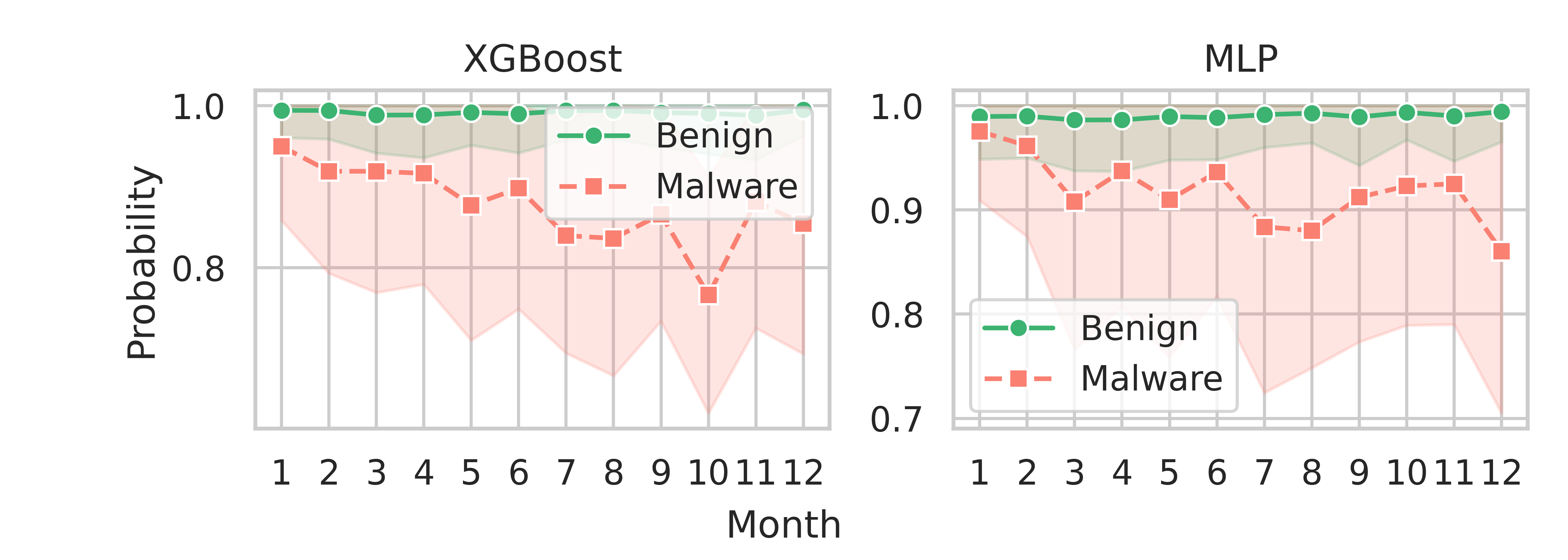}
        \caption{Drebin dataset}
        \label{fig:prob_month_drebin}
    \end{subfigure}
    
    \vspace{-0.05cm} 
    \begin{subfigure}[t]{0.99\linewidth}
        \centering
        \includegraphics[width=\linewidth]{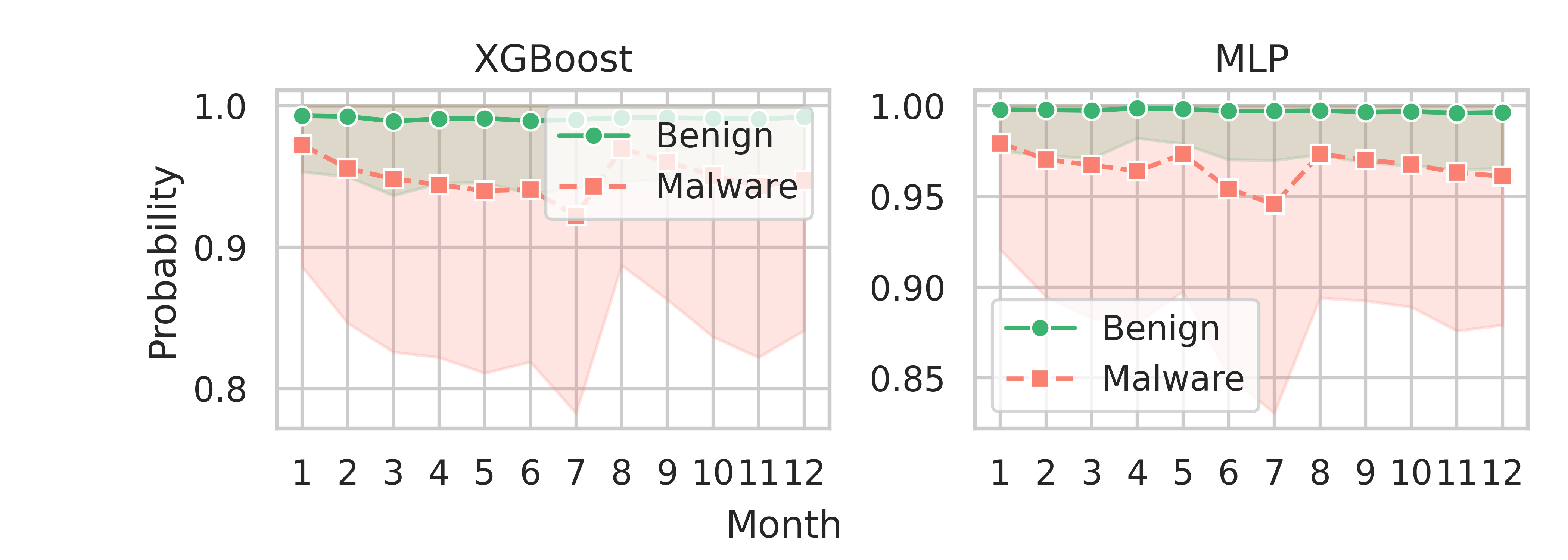}
        \caption{APIGraph dataset}
        \label{fig:prob_month_apigraph}
    \end{subfigure}

    \caption{Mean and standard deviation of predicted probabilities for benign and malware classes using XGBoost and MLP. Lower confidence for malware highlights that drift mainly affects this class.}

    \label{fig:prob_month_combined}
\end{figure}

Using a fixed thresholding approach for malware detection has limitations due to the divergent drift patterns in the benign and malware classes, as shown in Figure~\ref{fig:prob_month_combined}. In this figure, we plot the average prediction probability for the benign and malware classes using the XGBoost and MLP models on the Android datasets. The malware class experiences more severe drift, as indicated by its consistently lower average prediction probability than the benign class. 

To address this, we propose a class-dependent adaptive thresholding strategy for pseudo-label selection, which dynamically adjusts to the evolving data distribution. While dynamic thresholding has been applied in other domains~\cite{wang2024continual}, these approaches do not consider the distinct nature of drift across different classes. In our method, we define separate thresholds for the two classes: $\tau_m \in [0.5, 1]$ for malware and $\tau_b\in [0.5, 1]$ for benign samples. The adaptive thresholding mechanism adjusts these thresholds based on the model’s average prediction probabilities for each class over the unlabeled data. Specifically, we compute the mean probability for each class as follows:

\begin{align}
\mu_m &= \frac{1}{|\mathcal{D}_u^m|} \sum_{x_j \in \mathcal{D}_u^m} P(y = 1 \mid x_j) \label{eq:mu_m}\\
\mu_b &= \frac{1}{|\mathcal{D}_u^b|} \sum_{x_j \in \mathcal{D}_u^b} P(y = 0 \mid x_j) \label{eq:mu_b}
\end{align}

where $\mathcal{D}_u^m$ and $\mathcal{D}_u^b$ represent the subsets of the unlabeled dataset $\mathcal{D}_u$ that are predicted as malware and benign by the model $M$, respectively. The quantities  $\mu_m$ and $\mu_b$ are the mean probabilities for the malware and benign classes. While in our experimental setup, these values are computed using all the samples from the current month's data, the formulation can also be applied in a batch or online learning setting, where the quantities are computed based on batch statistics of the model’s predicted pseudo-labels.

We then combine these mean probabilities with the prior thresholds $\tau_m$ and $\tau_b$ using a parameter $\lambda \in [0, 1]$ to compute the updated thresholds, which accounts for the gradual shift in the data:

\begin{align}
\tau_m^{\text{updated}} &= \lambda \cdot \mu_m + (1 - \lambda) \cdot \tau_m \label{eq:tau_m_updated}\\
\tau_b^{\text{updated}} &= \lambda \cdot \mu_b + (1 - \lambda) \cdot \tau_b \label{eq:tau_b_updated}
\end{align}

After updating the thresholds, pseudo-labeled data is selected using these new adaptive thresholds. Specifically, if the model's predicted probability for a sample exceeds the updated threshold for malware ($\tau_m^{\text{updated}}$), the sample is assigned a label of 1 (malware). If the probability exceeds the updated threshold for benign ($\tau_b^{\text{updated}}$), it is labeled as 0 (benign). Otherwise, the sample remains unlabeled. Formally:

\[
\hat{y}_j = 
\begin{cases}
  1, & \text{if } P_M(y = 1 \mid x_j) > \tau_m^{\text{updated}} \\
  0, & \text{if } P_M(y = 0 \mid x_j) > \tau_b^{\text{updated}} \\
  \text{unlabeled}, & \text{otherwise}
\end{cases}
\]

The values of the threshold parameters $\tau_m$ and $\tau_b$, as well as the adaptation parameter $\lambda$, are jointly optimized during hyperparameter tuning, along with other model parameters, as discussed in Section~\ref{sec:hyperparameters}.

\subsection{Data Augmentation with Label Consistency}
\label{sec:augment}
Data augmentation is a commonly used technique in semi-supervised learning, particularly in contrastive learning, where the model is encouraged to produce consistent predictions for a sample and its perturbed counterpart~\cite{berthelot2019mixmatch,sohn2020fixmatch}. While data augmentation methods are prevalent in domains like computer vision, this is not the case for malware detection. In image classification, various label-preserving transformations, such as rotation, translation, scaling, and flipping, can be easily applied~\cite{cubuk2020randaugment}. However, designing equivalent transformations for malware datasets is significantly more challenging. Malware datasets typically consist of hand-crafted features, making it difficult to determine whether random transformations in the feature space preserve both the functionality and maliciousness of the application.

To address this challenge, Wu et al.~\cite{wu2023grim} proposed a data augmentation strategy for Windows malware classification in which features are randomly masked and replaced with features from other samples in the dataset. Given a sample $\mathbf{x} \in \mathbb{R}^d$, a mask vector $\mathbf{m} = [m_1, \dots, m_d]^\top \in \mathbb{R}^d$ is first generated, where each $m_j$ is independently sampled from a Bernoulli distribution with probability $p_a$. The augmented sample $\tilde{\mathbf{x}}$ is then obtained using the following equation:

\small
\begin{equation}
\tilde{\mathbf{x}} = (1 - \mathbf{m}) \odot \hat{\mathbf{x}} + \mathbf{x} \odot \mathbf{m}
\label{eq:eq_aug}
\end{equation}
\normalsize

where $\odot$ denotes element-wise multiplication, and $\hat{\mathbf{x}}$ is a new sample where each feature $\hat{x}_i$ is sampled from the empirical marginal distribution of the $i$-th feature. This empirical marginal distribution is represented by the uniform distribution over the values that the feature takes across the entire training dataset.

For example, consider a dataset with three features and four samples:
\[
\{[0, 1, 0], [1, 1, 0], [0, 0, 1], [0, 1, 1]\}.
\]
The value of the second feature of $\hat{\mathbf{x}}$ would be randomly sampled from the set $\{1, 1, 0, 1\}$, which are the observed values of the second feature across all samples.

\begin{figure}[t]
    \centering
    \includegraphics[width=\linewidth]{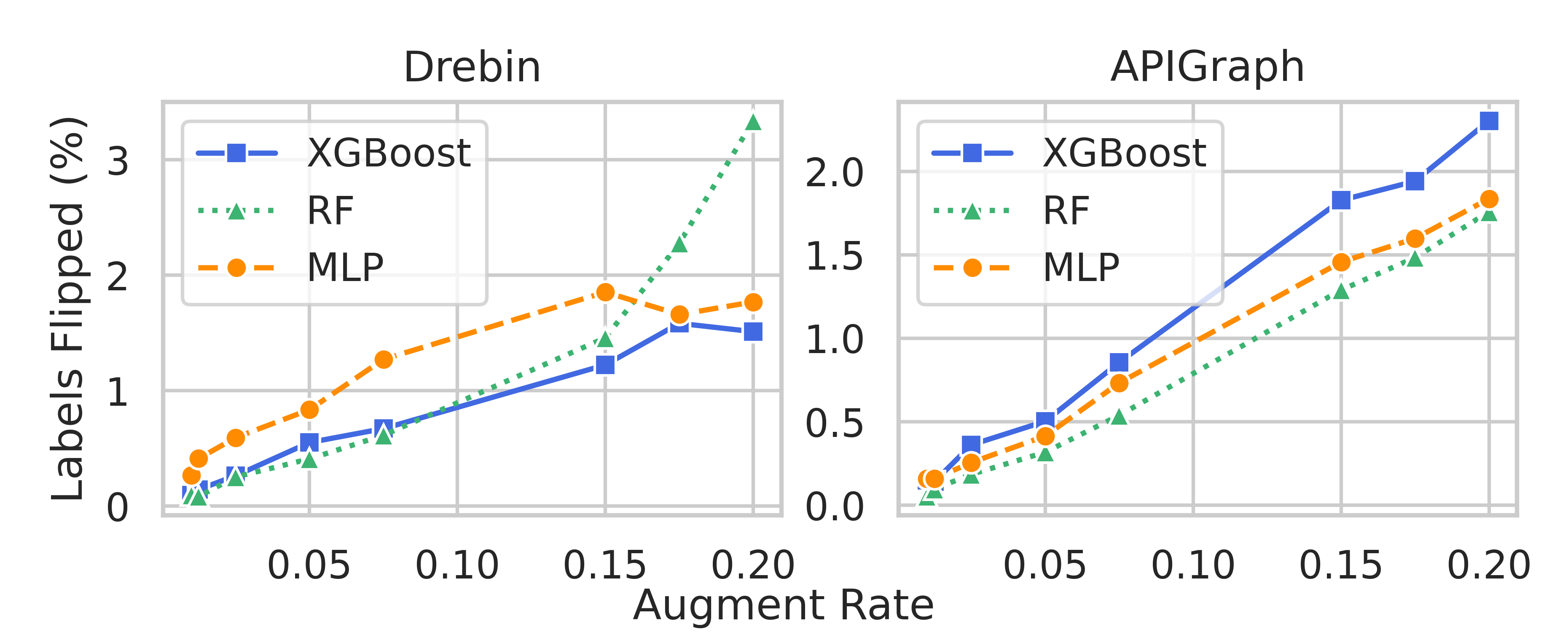}
    \caption{Percentage of Drebin and APIGraph samples where augmentation flips a correct prediction (i.e., benign to malicious or vice versa).
}
    \label{fig:augment_error}
\end{figure}

    
    


While the data augmentation strategy proposed by Wu et al.~\cite{wu2023grim} was designed for semi-supervised learning in a static setting where the model is trained only once, our scenario involves multiple model updates with additional pseudo-labeled data. Data augmentation can introduce samples whose labels may change due to concept drift in this setting. Such label flipping can degrade model performance in subsequent stages. We illustrate this effect in Figure~\ref{fig:augment_error} using the Drebin and APIGraph datasets across different models. Specifically, we select samples from the validation set where the model's original prediction is correct and apply augmentation with varying masking probabilities. We then recompute the model's predictions on the augmented data. Since augmentation should not alter the ground truth, if the model's prediction changes and becomes incorrect, we classify this as a case of label flipping. As the strength of the augmentation increases, the proportion of samples experiencing label flipping also rises. This suggests that overly strong augmentations can cause samples to cross the model's decision boundary, introducing additional noise into the learning process, which can accumulate over time. 

To address this, we propose two key modifications to ensure label consistency during augmentation. First, when replacing a masked feature, we restrict the replacement values only from samples of the same class. Although this setting was considered in~\cite{wu2023grim}, we argue that preserving label consistency is even more critical in our iterative training framework. Second, after the initial training phase, we leverage the current model to predict the label of the augmented sample. The augmented sample is included in the dataset only if the model's prediction matches the original label to improve label consistency. However, it is important to note that feature consistency is not explicitly enforced during augmentation. This means the generated features may not always represent valid samples, potentially leading to inconsistencies.

\subsection{Confidence Calibration with Mixup}
\label{sec:mixup}
One of the key challenges with self-training is the issue of confirmation bias (also referred to as self-poisoning), where the model reinforces its incorrect predictions~\cite{arazo2020pseudo}. This occurs when the model makes an incorrect prediction, uses that prediction to update itself, and consequently becomes more confident in that incorrect prediction in future iterations. 

Pseudo-labeling methods rely on thresholding to filter out low-confidence predictions and retain only high-confidence predictions. This process assumes that higher-confidence predictions are more likely to be correct. However, this assumption may not always hold due to issues with confidence calibration—the model's predicted confidence may not always reflect the true likelihood of the prediction being correct. By improving the model's confidence calibration, we can ensure that the pseudo-labels we incorporate into training are more accurate, thereby reducing the impact of self-poisoning.

To address this, we use \textit{mixup}\cite{mixup}, a regularization technique to improve model generalization and calibration. Mixup generates new training samples by interpolating both features and labels between two randomly selected samples. Specifically, given two samples, the features and labels are mixed based on a mixing coefficient $\lambda$ drawn from a Beta distribution parameterized by $\alpha$. This results in fractional labels when mixing samples from different classes. As shown in~\cite{thulasidasan2019mixup}, label interpolation is crucial for improving calibration. The target labels become fractional instead of one-hot, which changes the loss function for neural networks to the following:

\[
\mathcal{L}_{\text{mixup}} = \lambda \cdot \mathcal{L}(y_i, \hat{y}_i) + (1 - \lambda) \cdot \mathcal{L}(y_j, \hat{y}_i)
\]

where $\mathcal{L}$ is the binary cross-entropy loss, $y_i$ and $y_j$ are the true labels, and $\hat{y}_i$ is the model's prediction.

However, this approach is not applicable to models like Random Forests, which do not support fractional targets for classification tasks. To handle this, when mixing two samples with different labels, we assign the label corresponding to the sample with the higher interpolation coefficient. We provide evidence of how mixup improves confidence calibration in Section~\ref{sec:ablation}.

\section{Datasets} 
\label{sec:data}

\begin{table}[t]
\centering
\footnotesize
\caption{Summary statistics of the datasets}
\resizebox{0.95\columnwidth}{!}{%
\begin{tabular}{@{}ccccc@{}}

\toprule
\textbf{Dataset} & \textbf{Split} & \textbf{Duration} & \textbf{\begin{tabular}[c]{@{}c@{}}Benign \\ Apps\end{tabular}} & \textbf{\begin{tabular}[c]{@{}c@{}}Malicious\\ Apps\end{tabular}} \\ \toprule
\multirow{3}{*}{Drebin} & Train & 2019-01 to 2019-12 & 21449 & 2187 \\
 & Validation & 2020-01 to 2020-06 & 10549 & 777 \\
 & Test & 2020-07 to 2021-12 & 20106 & 822  \\ \midrule
\multirow{3}{*}{APIGraph} & Train & 2012-01 to 2012-12 & 16194 & 1447 \\
 & Validation & 2013-01 to 2013-06 & 13769 & 1446  \\
 & Test & 2013-07 to 2018-12 & 175048 & 13895  \\ \midrule
\multirow{3}{*}{BODMAS} & Train & 2019-10 to 2019-12 & 13763 & 11082 \\
 & Validation & 2020-01 to 2020-03 & 13206  & 13769 \\
 & Test & 2020-04 to 2020-09 & 32608   & 27701 \\ \midrule
\multirow{3}{*}{EMBER} & Train & 2018-01 &29423  & 32040 \\
 & Validation & -- & -- & -- \\
 & Test & 2018-02 to 2018-12 &320577  & 358864 \\ \midrule
\multirow{3}{*}{PDF} & Train & 10/13 to 10/27 &25233  & 1208 \\
 & Validation & 10/27 to 09/10 &70041 & 1893  \\
 & Test & 09/10 to 10/22 &135751  & 20224 \\ \bottomrule
\end{tabular}%
}

\label{tab:datasets}
\end{table}

We evaluate \texttt{ADAPT} across five publicly available malware detection datasets: two Android malware datasets, two Windows PE malware datasets, and one PDF malware dataset. Although the two Windows datasets share the same feature representation, they cover different time periods. The Android datasets differ both in feature sets and collection timelines. Table~\ref{tab:datasets} summarizes key statistics for each dataset.

\subsection{Android Datasets}
We use two Android malware detection datasets introduced by Chen et al.~\cite{chen2023continuous} for continuous learning using static features. The first dataset utilizes the Drebin feature set~\cite{arp2014drebin}, while the second employs the APIGraph feature set~\cite{zhang2020enhancing}. 



The Drebin and APIGraph datasets represent Android applications using binary feature vectors that indicate the presence or absence of specific attributes. Drebin includes 16,978 dimensional features spanning eight categories: permissions, intents, API usage, and network addresses. In contrast, APIGraph reduces the feature space to 1,159 dimensions by clustering semantically similar APIs. Drebin samples were collected between 2019 and 2021, while APIGraph spans from 2012 to 2018. Both datasets were curated to mitigate experimental bias in malware detection~\cite{pendlebury2019tesseract}, though they originally contained duplicates, which can affect result consistency~\cite{zhao2021impact,alam2024revisiting}. We, therefore, use deduplicated versions of both datasets in our experiments.

For both datasets, we use the first year's data as the training set, the next six months for validation, and the remaining months for testing. For pseudo-labeling, the training data is used to train the initial model, which is then periodically updated with data from subsequent months in the validation and test sets. The validation set is also used to jointly tune the model's hyperparameters and those specific to the pseudo-labeling algorithm.

\subsection{Windows Datasets}
We use two Windows PE malware datasets: EMBER~\cite{2018arXiv180404637A} and BODMAS~\cite{yang2021bodmas}. Both datasets share identical feature sets but span different periods. Each file's features, extracted using the LIEF project~\cite{LIEF}, consist of byte sequences, imported functions, and header information compiled into a 2,381-dimensional feature vector.

Since both datasets have the same feature representation, we perform hyperparameter tuning only on the BODMAS dataset and use those tuned hyperparameters to evaluate the EMBER dataset. For the BODMAS dataset, we use the first three months of data for training, the next three months for validation, and the remaining six months for testing. For the EMBER dataset, we treat the first month's data as the labeled training set and use the remaining months for testing.

\subsection{PDF Dataset}
We use the PDF malware dataset from~\cite{vsrndic2016hidost} in our study. The dataset contains features representing benign and malicious PDFs over ten weeks, from 09/13/2012 to 10/22/2012. Although the number of features varies between weeks, we restrict our analysis to the features consistently present across all weeks to ensure compatibility with the model architectures used in our experiments. This results in 950-dimensional feature vectors. Compared to the Android and Windows datasets, this dataset exhibits less concept drift due to the shorter period. 

For training, we randomly sample 10\% of the data from the first two weeks. The following two weeks are used for validation, and the test set consists of data from the final six weeks.

\section{Experimental Settings}

\begin{table*}[t]
\caption{
Comparison of methods across all datasets. Results are the mean of five runs (five different random seeds). Rows labeled \textbf{Oracle} represent an ideal scenario where retraining uses ground-truth labels; these are provided for reference only and are not considered in best result highlighting. For each dataset, \underline{underlined} F1-scores indicate the best baseline (excluding proposed and Oracle methods), and \textbf{\underline{bold+underline}} values indicate the best overall result (including proposed methods).
}

\centering
\adjustbox{max width=\textwidth}{
\begin{tabular}{ll|ccc|ccc|ccc|ccc|ccc}
    \toprule
    & & \multicolumn{3}{c|}{\textbf{Drebin}} & \multicolumn{3}{c|}{\textbf{APIGraph}} & \multicolumn{3}{c|}{\textbf{BODMAS}} & \multicolumn{3}{c|}{\textbf{EMBER}} & \multicolumn{3}{c}{\textbf{PDF Malware}} \\
    \cmidrule(lr){3-5}\cmidrule(lr){6-8}\cmidrule(lr){9-11}\cmidrule(lr){12-14}\cmidrule(lr){15-17}
    \textbf{Method} & \textbf{Model} & F1 & FPR & FNR & F1 & FPR & FNR & F1 & FPR & FNR & F1 & FPR & FNR & F1 & FPR & FNR \\
    \midrule

    \multirow{3}{*}{Offline ML} 
        & RF      & 33.2 & 0.14 & 78.7
                & 46.8 & 0.34 & 66.9
                & 96.3 & 0.21 & 6.69
                & 86.3 & 3.47 & 21.7
                & \underline{98.0} & 0.07 & 3.31 \\
        & XGBoost & 44.9 & 0.66 & 65.6
                & 66.5 & 0.97 & 42.7
                & \underline{99.2} & 0.30 & 1.23
                & \underline{92.3} & 3.58 & 11.4
                & 97.9 & 0.14 & 3.17 \\
        & MLP     & 40.9 & 0.57 & 69.8
                & 57.0 & 1.67 & 49.9
                & 98.2 & 0.64 & 2.89
                & 88.1 & 5.66 & 17.1
                & 97.4 & 0.11 & 4.26 \\
    \midrule

        \multirow{3}{*}{Oracle} 
        & RF      & 53.4 & 0.12 & 61.3
                & 78.9 & 0.24 & 31.8
                & 96.6 & 0.19 & 6.22
                & 87.1 & 3.38 & 20.4
                & 98.8 & 0.08 & 1.85 \\
        & XGBoost & 68.3 & 0.52 & 41.2
                & 85.5 & 0.67 & 17.6
                & 99.6 & 0.21 & 0.61
                & 93.1 & 3.46 & 10.1
                & 99.2 & 0.09 & 1.04 \\
        & MLP     & 74.6 & 0.53 & 33.1
                & 89.0 & 0.45 & 14.4
                & 98.8 & 0.50 & 1.77
                & 90.0 & 5.16 & 14.2
                & 97.6 & 0.09 & 3.87 \\
    \midrule

    \multirow{4}{*}{Self-Training}
        & ARF      & 34.0 & 9.88 & 45.8
                & 61.1 & 5.28 & 25.0
                & 97.5 & 2.13 & 2.28
                & 69.7 & 91.8 & 1.78
                & 29.7 & 0.01 & 80.0 \\
        & DE++     & 34.7 & 7.84 & 47.1
                & 42.3 & 0.33 & 65.6
                & 90.8 & 15.1 & 1.83
                & 76.9 & 24.9 & 22.8
                & 60.8 & 19.3 & 0.28 \\
        & Insomnia & \underline{49.8} & 0.70 & 60.6
                & 68.1 & 2.05 & 31.8
                & 97.2 & 0.56 & 4.81
                & 87.1 & 6.06 & 18.5
                & 88.2 & 0.09 & 15.4 \\
        & MORSE    & 41.2 & 4.50 & 54.7
                & \underline{75.3} & 1.82 & 22.2
                & 98.8 & 0.48 & 1.73
                & 89.8 & 6.34 & 13.7
                & 97.4 & 0.12 & 4.16 \\
    \midrule

    \multirow{3}{*}{Proposed}
        & RF+\texttt{ADAPT} & 49.8 & 0.21 & 64.3
                & 67.2 & 0.34 & 46.0
                & 96.1 & 0.22 & 7.11
                & 86.2 & 3.56 & 21.7
                & \textbf{\underline{99.1}} & 0.09 & 1.26 \\
        & XGBoost+\texttt{ADAPT} & \textbf{\underline{57.1}} & 0.24 & 56.6
                & 73.2 & 2.26 & 22.2
                & \textbf{\underline{99.6}} & 0.24 & 0.61
                & \textbf{\underline{93.5}} & 4.45 & 8.70
                & 98.6 & 0.10 & 2.03 \\
        & MLP+\texttt{ADAPT} & 50.5 & 0.77 & 59.1
                & \textbf{\underline{76.8}} & 0.81 & 29.4
                & 98.6 & 0.41 & 2.26
                & 89.1 & 5.01 & 16.0
                & 97.2 & 0.09 & 4.73 \\

    \bottomrule
\end{tabular}
}
\label{tab:all-results}
\end{table*}

\subsection{Baseline Machine Learning Models}


Our proposed algorithm is compatible with most machine learning models, requiring only access to model probabilities for pseudo-labeling, with data augmentation applied in a model-agnostic manner. We evaluate its performance using three widely adopted baseline models: Random Forest (RF)~\cite{breiman2001random}, XGBoost~\cite{chen2016xgboost}, and a Multi-Layer Perceptron (MLP)~\cite{rumelhart1986learning}, chosen for their relevance in malware classification tasks~\cite{alam2013random, dambra2023decoding, pendlebury2019tesseract, zhang2020enhancing, 2018arXiv180404637A, yang2021bodmas}. RF is computationally efficient, interpretable, and performs well on smaller datasets. XGBoost excels on tabular data, typical in malware detection, due to its ability to capture complex feature interactions~\cite{grinsztajn2022tree, alam2024revisiting}. MLPs are well-suited for large-scale semi-supervised settings, supporting batch training and dynamic data augmentation, and exhibit robustness to noisy labels~\cite{wu2023grim}, making them ideal for pseudo-labeling methods~\cite{sohn2020fixmatch}.

\subsection{Self-Training Baselines}

We compare our method against four self-training baselines: Adaptive Random Forest (ARF)~\cite{gomes2017adaptive}, DroidEvolver++ (DE++)~\cite{kan2021investigating}, Insomnia~\cite{andresini2021insomnia}, and MORSE~\cite{wu2023grim}. ARF and DE++ are online learning methods that address concept drift without requiring access to the original training dataset. In contrast, Insomnia and MORSE follow a semi-supervised paradigm, leveraging the original training data and pseudo-labeled samples for model updates. While our approach (\texttt{ADAPT}) also assumes access to the original training data, we additionally explore scenarios where this access is restricted. Further implementation details for all baseline methods are provided in Appendix~\ref{app:st}, and the restricted setting without original training data is discussed in Appendix~\ref{sec:sf-adapt}.

\subsection{Hyperparameter Tuning}
\label{sec:hyperparameters}

As shown in previous work~\cite{chen2023continuous,alam2024revisiting}, continuous learning settings require dedicated hyperparameter tuning. In our experiments, we jointly tune the hyperparameters of both the baseline models and our proposed approach using the validation set. \texttt{ADAPT} introduces five key hyperparameters: the benign base threshold ($\tau_b$), the malware base threshold ($\tau_m$), the adaptive threshold weight ($\lambda$), the masking probability for augmentation ($p_a$), and the mixup parameter ($\alpha$). Each baseline self-training method also has its hyperparameters; for example, ARF uses a threshold ($\tau$), DE++ tunes the app buffer size, malware-to-benign ratio, and model aging threshold, Insomnia adjusts the pseudo-labeling fraction, and MORSE tunes the threshold along with weak and strong augmentation probabilities. A complete list of hyperparameters and their ranges is provided in Appendix~\ref{sec-app--hpo}. To ensure fair comparison despite varying hyperparameter counts, we follow~\cite{alam2024revisiting} and use a fixed budget of 200 random hyperparameter searches~\cite{bergstra2012random} for each method on each dataset.

\subsection{Evaluation}
We use the same performance metrics as in~\cite{chen2023continuous}: the average F1-score, False Positive Rate (FPR), and False Negative Rate (FNR) across all months. Experiments are conducted using the optimal hyperparameters, with five random seeds on the test set. We report the mean and standard deviation of the performance metrics across these runs.
\section{Results and Analysis}

\begin{figure*}[t]
    \centering
    \includegraphics[width=\linewidth]{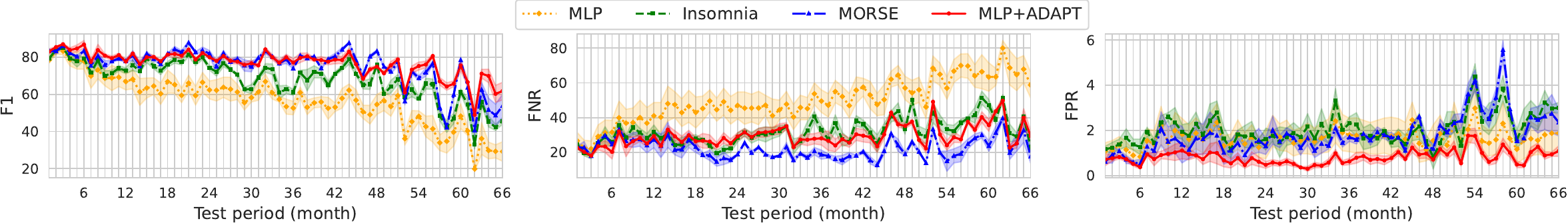}
    \caption{F1-score over test months on APIGraph dataset.}
    \label{fig:result-apigraph}
\end{figure*}

\begin{figure}[t]
    \centering
    \includegraphics[width=\linewidth]{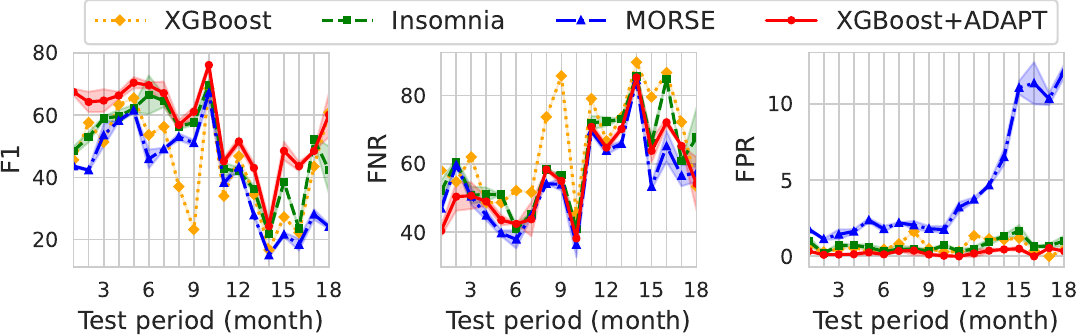}
    \caption{Performance on the Drebin dataset: F1 score (left), FNR (middle), and FPR (right) over test months.}
    \label{fig:result-drebin}
\end{figure}

\begin{figure}[t]
    \centering
    \includegraphics[width=\linewidth]{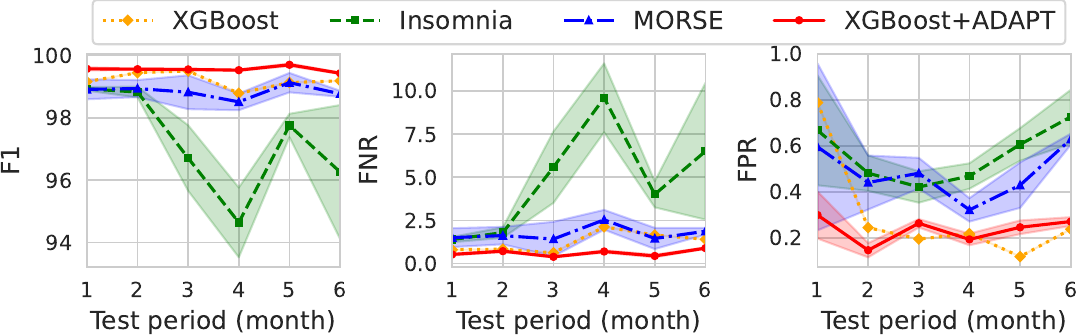}
    \caption{Performance on the BODMAS dataset: F1 score (left), FNR (middle), and FPR (right) over test months.}
    \label{fig:result-bodmas}
\end{figure}

\begin{figure}[t]
    \centering
    \includegraphics[width=\linewidth]{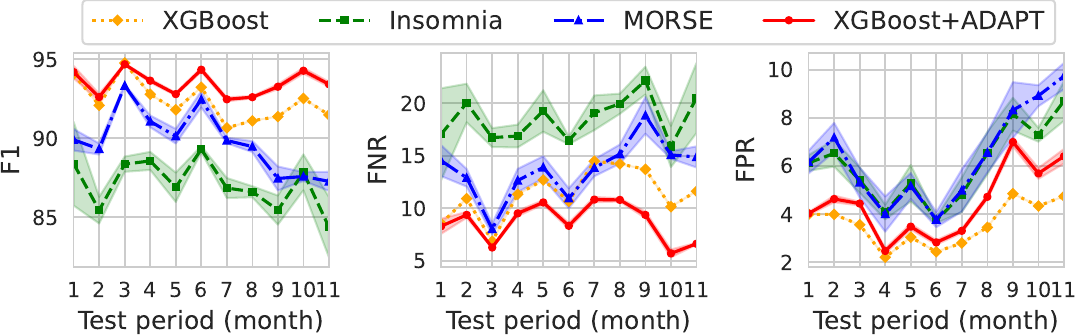}
    \caption{Performance on the EMBER dataset: F1 score (left), FNR (middle), and FPR (right) over test months.}
    \label{fig:result-ember}
\end{figure}
\begin{figure}[t]
    \centering
    \includegraphics[width=\linewidth]{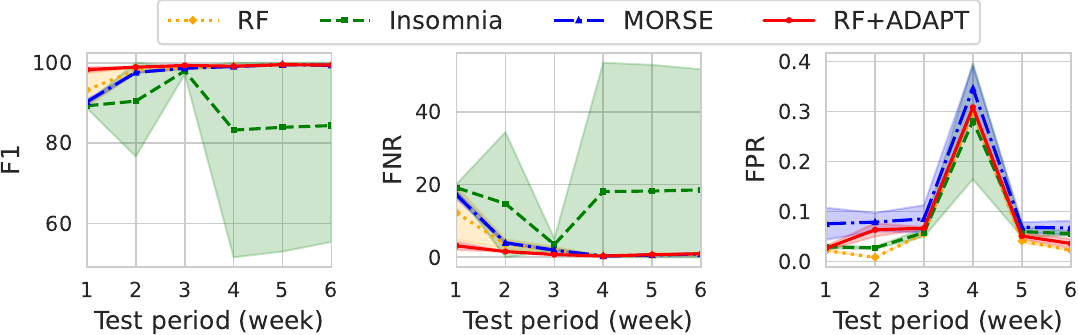}
    \caption{Performance on the PDF dataset: F1 score (left), FNR (middle), and FPR (right) over test months.}
    \label{fig:result-pdf}
\end{figure}

Table~\ref{tab:all-results} presents the results of various methods across five datasets. XGBoost+\texttt{ADAPT} achieves the highest overall F1-score on the Drebin, BODMAS, and EMBER datasets, while MLP+\texttt{ADAPT} attains the best performance on the APIGraph dataset. Random Forest+\texttt{ADAPT} yields the top result on the PDF malware dataset. Across all benchmarks, \texttt{ADAPT} consistently enhances the performance of baseline machine learning models, with particularly notable gains on the Android malware dataset, where concept drift causes greater performance degradation.

We also report results for an \textbf{Oracle} setting, where ground-truth labels are available for retraining, but augmentation and mixup components used in \texttt{ADAPT} are omitted. The results indicate that, for the Drebin and APIGraph datasets, the F1-score of \texttt{ADAPT} lags behind the Oracle model by more than 10\%. However, this gap is less pronounced on the other three datasets, and on the EMBER dataset, XGBoost + \texttt{ADAPT} even surpasses the Oracle, likely due to the benefits introduced by data augmentation and mixup.

    

\paragraph*{\textbf{Drebin Dataset}}

Drebin is the most challenging dataset, as evidenced by the consistently low F1-scores of the offline machine learning models. Among the four self-training baselines, ARF, DE++, and MORSE do not yield notable improvements over the offline models in terms of F1-score. In contrast, Insomnia outperforms the offline MLP and XGBoost models, highlighting the effectiveness of our Insomnia adaptation for malware detection under concept drift. \texttt{ADAPT} achieves substantial improvements across all baseline models, with absolute F1-score gains of 16.6\%, 12.2\%, and 9.6\% for RF, XGBoost, and MLP, respectively. This improvement is primarily attributed to a significant reduction in false negatives. Notably, our method does not increase the false positive rate; in fact, it reduces the false positive rate by 63.6\% compared to the baseline XGBoost model.

Figure~\ref{fig:result-drebin} illustrates the temporal dynamics of F1, FPR, and FNR over the test months for four models on the Drebin dataset. Notably, the XGBoost+\texttt{ADAPT} method consistently outperforms the other models in F1-score across most test months and is particularly effective at mitigating the sharp performance drop experienced by the baseline XGBoost model in certain months, such as month 9 (2021-03).

    

\paragraph*{\textbf{APIGraph Dataset}}
The baseline machine learning models achieve higher performance on the APIGraph dataset compared to Drebin, as its features are specifically designed to be more resilient to concept drift~\cite{zhang2020enhancing}. Among the self-training baselines, all except DE++ show improvements over the offline models. Our proposed \texttt{ADAPT} method further improves upon the baselines, particularly for RF and MLP, both of which exhibit an approximately 20\% increase in F1-score. While MORSE achieves the lowest False Negative Rate, the MLP+\texttt{ADAPT} method attains a lower False Positive Rate along with the best overall F1-score.

Figure~\ref{fig:result-apigraph} illustrates the F1-score over the test months for four models on the APIGraph dataset. The performance of all three adaptation models remains relatively stable during the first two years of test data, after which Insomnia experiences a sharp decline. \texttt{ADAPT} and MORSE perform similarly throughout most of the test period, except towards the end where more significant drift occurs.

\paragraph*{\textbf{BODMAS Dataset}}

The baseline models achieve notably high F1-scores on the BODMAS dataset, indicating that concept drift has a less pronounced effect in this setting. None of the self-training baseline methods surpasses the performance of the baseline XGBoost model, which already attains an F1-score of 99.2\%. Our proposed \texttt{ADAPT} method further increases this score by 0.4\%, while also reducing both the False Positive Rate (FPR) and False Negative Rate (FNR). This result underscores the value of our model-agnostic approach, as tree-based models such as XGBoost may inherently outperform neural networks on certain datasets~\cite{grinsztajn2022tree}. A similar improvement is observed for MLP, although the performance of RF decreases. Figure~\ref{fig:result-bodmas} presents the monthly results for different models, highlighting that XGBoost+\texttt{ADAPT} consistently maintains high performance throughout the test period.

\begin{figure*}[t]
    \centering
    \includegraphics[width=\linewidth]{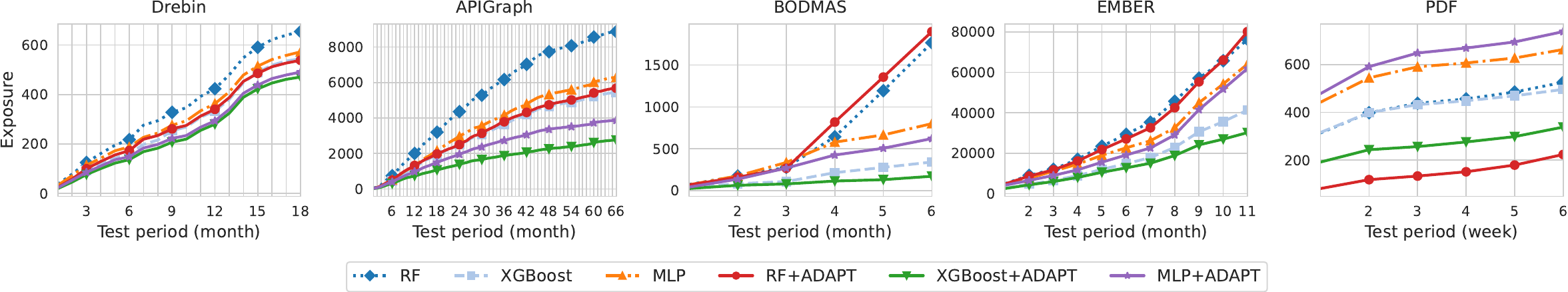}
    \caption{Exposure over test months on different datasets.}
    \label{fig:exposure}
\end{figure*}

\paragraph*{\textbf{EMBER Dataset}}
Performance on the EMBER dataset is slightly lower, mainly due to the use of only one month of training data and the absence of separate hyperparameter tuning, in contrast to the three months of training used for BODMAS. On EMBER, the baseline XGBoost model outperforms all self-training methods, but our proposed \texttt{ADAPT} method still achieves a further 1.2\% improvement in F1-score over the XGBoost baseline. Among the self-training baselines, DE++ and ARF perform significantly worse than the offline models, indicating greater susceptibility to self-poisoning and a high sensitivity to threshold settings in this dataset. Figure~\ref{fig:result-ember} shows the monthly performance of different models, highlighting that XGBoost+\texttt{ADAPT} consistently achieves the highest performance throughout the test period.

    

\paragraph*{\textbf{PDF Dataset}}

Figure~\ref{fig:result-pdf} shows the weekly performance of different models on the PDF malware dataset. Random Forest outperforms both XGBoost and MLP, suggesting that more complex models may be susceptible to overfitting in this setting. Nonetheless, our proposed method, \texttt{ADAPT}, improves the F1-score by 1.1\%, primarily through a reduction in the False Negative Rate (FNR).

\subsection{Exposure Analysis}

To better capture the real-world risk associated with missed detections, we evaluate models using the Absolute Exposure (AE) metric introduced in~\cite{botacin2025towards}. AE quantifies the cumulative number of false negatives over time, reflecting the duration and extent to which users remain exposed to undetected threats. Unlike traditional accuracy metrics, AE directly measures how long malicious samples evade detection, providing a more practical assessment of security risk in streaming or temporally evolving scenarios.

Figure~\ref{fig:exposure} presents the Absolute Exposure values for baseline and \texttt{ADAPT} models across all datasets. Notably, XGBoost+\texttt{ADAPT} achieves the lowest AE in every dataset except for the PDF malware dataset, where Random Forest + \texttt{ADAPT} attains the best performance.

\begin{figure}[t]
    \centering
    \includegraphics[width=\linewidth]{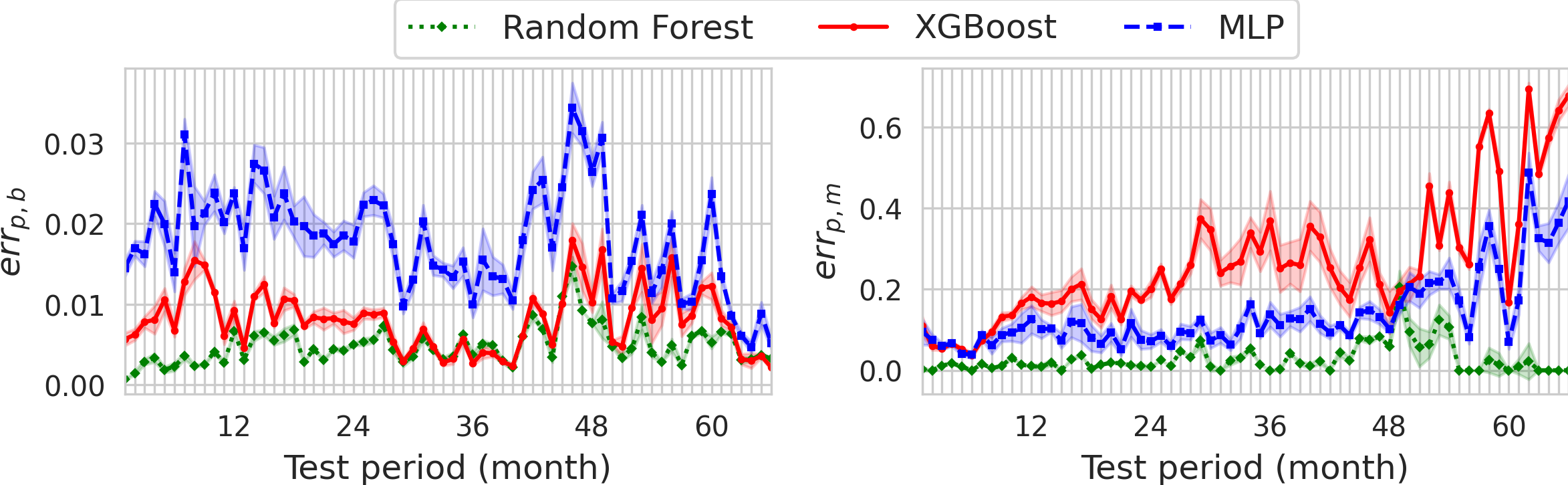}
    \caption{Pseudo-labeling errors during the test months on the APIGraph dataset. (Left): Fraction of samples incorrectly pseudo-labeled as benign. (Right): Fraction of samples incorrectly pseudo-labeled as malware.}
    \label{fig:pl-error}
\end{figure}

\subsection{Effectiveness of ADAPT Across Models}
We designed \texttt{ADAPT} as a model-agnostic adaptation strategy and demonstrated its ability to improve performance across different models. Our goal was to show that ADAPT enhances baseline models (RF, XGBoost, MLP) in the presence of concept drift. While different models may be preferable in various deployment settings, XGBoost+\texttt{ADAPT} emerges as a strong default choice, achieving the highest F1-score on three out of five datasets.  

Although MLP+\texttt{ADAPT} achieves the highest performance over the full six-year test period on the APIGraph dataset, XGBoost+\texttt{ADAPT} (83.7\% F1) outperforms both MLP+\texttt{ADAPT} (82.7\% F1) and MORSE (80.5\% F1) in the first year of test data, a timeframe that more realistically represents practical deployment scenarios for unlabeled adaptation. This highlights that while ADAPT is model-agnostic, its strong synergy with XGBoost enhances its versatility as a broadly applicable solution. The statistical analysis of \texttt{ADAPT}, provided in Section~\ref{stat-sign} further substantiates this finding.

\subsection{Error Analysis}
\label{sec-err}

Table~\ref{err-pdf} summarizes the changes in False Positive Rate (FPR) and False Negative Rate (FNR) between baseline models and those trained with \texttt{ADAPT}. Across datasets, F1-score improvements are mainly due to reduced FNR, except for Random Forest on BODMAS and MLP on PDF. This suggests that \texttt{ADAPT} enables the model to better learn from uncertain malware samples, improving its ability to correctly classify similar cases upon retraining. In most scenarios, FPR is maintained or reduced alongside FNR, though exceptions exist—e.g., XGBoost on APIGraph, which shows decreased FNR but increased FPR. Neural networks generally outperform other models in reducing FPR. We further examine this in the context of pseudo-labeling errors on APIGraph.

Figure~\ref{fig:pl-error} shows the pseudo-labeling errors introduced during the test months for the three \texttt{ADAPT}-trained models on the APIGraph dataset. Here, $err_{p,b}$ denotes the fraction of malware samples incorrectly pseudo-labeled as benign, and $err_{p,m}$ denotes benign samples mislabeled as malware. Across all models, $err_{p,b}$ is significantly lower than $err_{p,m}$, indicating that benign pseudo-labels are generally more accurate. This can be attributed to the higher thresholds for selecting benign samples: the optimal values of $\tau_b$ for Random Forest, XGBoost, and MLP are 0.96, 0.97, and 0.98, respectively. These high thresholds ensure that only samples with high benign confidence are selected, which helps prevent FNR degradation. This strategy is essential, as evasive malware may exhibit features similar to benign samples. Lowering the threshold could result in mislabeling such malware as benign, negatively impacting FNR.

Conversely, $err_{p,m}$ is notably higher due to the lower threshold $\tau_m$ used for pseudo-labeling malware samples—0.67, 0.70, and 0.63 for Random Forest, XGBoost, and MLP, respectively. These lower thresholds increase the likelihood of mislabeling benign samples as malware, a problem that intensifies over time, especially beyond four years. This explains the rise in FPR for XGBoost. Despite also introducing many noisy pseudo-labels, the MLP model reduces FPR compared to its baseline, likely due to the inherent robustness of neural networks to label noise in semi-supervised learning with data augmentation~\cite{wu2023grim}. From a threat modeling perspective, using a lower threshold for malware selection is reasonable—benign apps lack the motivation to evade detection, so lower-confidence malware predictions still provide valuable drifted samples for retraining. While this increases the risk of labeling errors, it is essential for maintaining adaptability against evolving threats.



\begin{table}[t]
\caption{Difference between False Positive Rate (FPR) and False Negative Rate (FNR) across different models and datasets. Positive values indicate a reduction in error, while negative values indicate an increase in error using the \texttt{ADAPT}.}
    \centering
    \resizebox{\columnwidth}{!}{%
    \begin{tabular}{l|l|ccccc}  
        \toprule
        \textbf{Model} & \textbf{Metric} & \textbf{Drebin} & \textbf{APIGraph} & \textbf{BODMAS} & \textbf{EMBER} & \textbf{PDF} \\
        \midrule
        \multirow{2}{*}{RF} & $\Delta$FPR & -0.07 & 0.00 & -0.01 & -0.09 & -0.02 \\
                                      & $\Delta$FNR & 14.4 & 20.9 & -0.42 & 0.00 & 2.05 \\
        \midrule
        \multirow{2}{*}{XGBoost} & $\Delta$FPR & 0.42 & -1.30 & 0.06 & -0.87 & 0.04 \\
                                          & $\Delta$FNR & 9.00 & 20.5 & 0.62 & 2.70 & 1.14 \\
        \midrule
        \multirow{2}{*}{MLP} & $\Delta$FPR & -0.20 & 0.86 & 0.23 & 0.65 & 0.02 \\
                                       & $\Delta$FNR & 10.7 & 20.5 & 0.63 & 1.10 & -0.47 \\
        \bottomrule
    \end{tabular}
    }
    
    \label{err-pdf}
\end{table}

\begin{table}[t]
\caption{Performance metrics (F1, FPR, FNR) for RF, XGBoost, and MLP with various components removed on the first six months of the APIGraph test set.}
\centering
\resizebox{\columnwidth}{!}{
\begin{tabular}{l|ccc|ccc|ccc}
\toprule
\multirow{2}{*}{\makecell{Component \\ Removed}} & \multicolumn{3}{c|}{RF} & \multicolumn{3}{c|}{XGBoost} & \multicolumn{3}{c}{MLP} \\
\cmidrule(r){2-4} \cmidrule(r){5-7} \cmidrule(r){8-10}
 & F1 & FPR & FNR & F1 & FPR & FNR & F1 & FPR & FNR \\
\midrule
None                & 83.6 & 0.25 & 26.6 & 86.7 & 1.06 & 16.0 & 85.2 & 0.66 & 21.4 \\
Adaptive Th.        & 80.8 & 0.17 & 31.2 & 86.4 & 1.00 & 16.8 & 81.2 & 0.50 & 25.1 \\
Augmentation        & 82.3 & 0.21 & 28.8 & 85.6 & 1.11 & 17.5 & 84.8 & 0.75 & 21.4 \\
Mixup               & 83.5 & 0.25 & 26.7 & 86.5 & 0.97 & 16.9 & 81.5 & 0.54 & 27.8 \\
\bottomrule
\end{tabular}
 
}

\label{tab-abl}
\end{table}

\subsection{Ablation and Mechanism Analysis}
\label{sec:ablation}

We conduct an ablation study by removing different components of \texttt{ADAPT} on the APIGraph dataset, as shown in Table~\ref{tab-abl} for the first six months of test data. In this experiment, we replace the class-specific adaptive thresholding scheme with a fixed threshold of $\tau_b$. We observe that each component contributes to the overall improvement in F1-score and FNR, albeit varying degrees across different models. While replacing adaptive thresholding reduces the FPR marginally due to the higher threshold, as discussed in Section~\ref{sec-err}, it leads to a significant increase in the FNR, as drifted malware samples are not included in the retraining phase.

\begin{table}[t]
\caption{Feature Correlations Before and After Augmentation}
\centering
\begin{tabular}{lcc}
\toprule
\textbf{Dataset} & \textbf{Mean $|\rho(X)|$} & \textbf{Spearman ($\rho(X)$, $\rho(X_{\text{aug}})$)} \\
\midrule
APIGraph & 0.0238 & 0.9247 \\
Drebin   & 0.0269 & 0.9911 \\
Bodmas   & 0.0276 & 0.9491 \\
PDF      & 0.0857 & 0.9975 \\
\bottomrule
\end{tabular}
\label{tab:corr_comparison}
\end{table}

\begin{figure}[t]
    \centering
        \centering    \includegraphics[width=\linewidth]{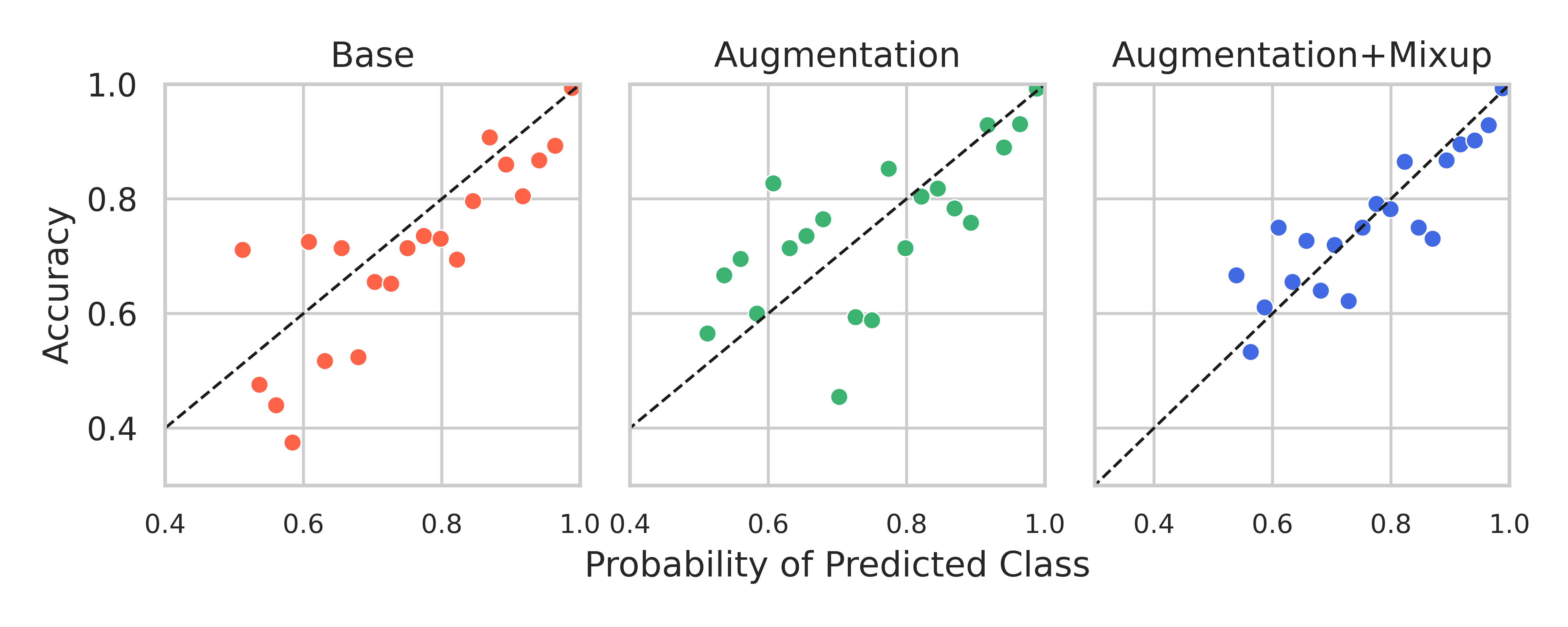}
        \label{fig:prob_month_apigraph}
    \caption{Calibration plots for the XGBoost model on the APIGraph dataset. The model trained with Mixup shows better calibration, as the points are lying closer to the $y=x$ line.}
    \label{fig:calibration}
\end{figure}

\textbf{Data Augmentation \& Consistency.} Our feature-space augmentation does not guarantee valid samples, as modeling complex feature dependencies is intractable. To assess the impact, we analyze whether augmentation disrupts feature relationships by computing the mean absolute correlation in the original data and the Spearman correlation between original and augmented correlation matrices. As shown in Table~\ref{tab:corr_comparison}, low mean correlations indicate weak dependencies, and high Spearman values confirm that feature relationships are largely preserved, suggesting that the overall statistical structure remains stable.

\textbf{Effect of Mixup.} Figure~\ref{fig:calibration} shows the effect of Mixup on calibration for XGBoost on APIGraph. The scatter plots compare accuracy versus confidence for three models: base (no augmentation or Mixup), with augmentation only, and with both augmentation and Mixup. The base model is notably overconfident, while data augmentation improves calibration, and combining it with Mixup provides the closest alignment between confidence and accuracy, especially for high-confidence predictions.

\subsection{Statistical Significance}
\label{stat-sign}
Table \ref{tab:wilcoxon_results} presents the Wilcoxon signed-rank test p-values for comparing different machine learning models (Random Forest, MLP, and XGBoost) across multiple malware datasets. The Wilcoxon test~\cite{wilcoxon1992individual} is a non-parametric statistical test used to assess whether two paired samples come from the same distribution. In this case, it evaluates whether the adapted model's F1-scores significantly differ from the baseline in the test months. Cells are shaded in gray if the p-value is less than 0.05, indicating statistically significant differences. Notably, we observe consistently low p-values for the XGBoost models across all datasets, suggesting a significant effect of adaptation on their performance.

\begin{table}[h]
\caption{Wilcoxon $p$-values for different models across datasets. Cells with $p < 0.05$ are shaded.}
\centering
\resizebox{\columnwidth}{!}{%
\setlength{\tabcolsep}{12pt}
\begin{tabular}{lccc}
    \toprule
    \textbf{Dataset} & \textbf{Random Forest} & \textbf{MLP} & \textbf{XGBoost} \\
    \midrule
    Drebin    & \cellcolor{gray!30} $7.63 \times 10^{-6}$  & \cellcolor{gray!30} $5.34 \times 10^{-5}$  & \cellcolor{gray!30} $1.53 \times 10^{-5}$  \\
    APIGraph  & \cellcolor{gray!30} $1.64 \times 10^{-12}$ & \cellcolor{gray!30} $1.64 \times 10^{-12}$ & \cellcolor{gray!30} $1.23 \times 10^{-10}$ \\
    BODMAS    & $0.6875$                                 & \cellcolor{gray!30} $0.03125$               & \cellcolor{gray!30} $0.03125$              \\
    EMBER     & $0.7646$                                 & \cellcolor{gray!30} $0.03223$               & \cellcolor{gray!30} $0.00195$              \\
    PDF       & $0.4375$                                 & $0.15625$                                   & \cellcolor{gray!30} $0.03125$              \\
    \bottomrule
\end{tabular}
}
\label{tab:wilcoxon_results}
\end{table}

\begin{table}[t]
\caption{Performance on Drebin and APIGraph with active learning (AL) with 50 monthly annotation budget.}
    \centering
    \resizebox{\columnwidth}{!}{%
    \begin{tabular}{ll|ccc}
        \toprule
        \textbf{Dataset} & \textbf{Method} & \textbf{F1} & \textbf{FPR} & \textbf{FNR} \\
        \midrule
        \multirow{4}{*}{Drebin} 
        & XGBoost & 62.5±1.95 & 0.57±0.04 & 47.7±1.85 \\
        & XGBoost + AL & \underline{81.3±0.00} & 0.14±0.00 & 29.0±0.00 \\
        & HCC & 68.5±3.38 & 0.57±0.05 & 40.3±4.37 \\
        & XGBoost + \texttt{ADAPT} + AL & \textbf{\underline{81.7±0.28}} & 0.16±0.01 & 27.8±0.44 \\
        \midrule
        \multirow{4}{*}{APIGraph} 
        & XGBoost & 79.2±0.43 & 0.98±0.06 & 24.2±1.20 \\
        & XGBoost + AL & \underline{88.7±0.00} & 0.36±0.00 & 15.8±0.00 \\
        & HCC  & 86.7±0.15 & 0.56±0.03 & 17.2±0.36 \\
        & XGBoost + \texttt{ADAPT} + AL & \textbf{\underline{89.4±0.17}} & 0.79±0.02 & 9.55±0.18 \\
        \bottomrule
    \end{tabular}%
    }
    
    \label{tab:res-al}
\end{table}

\section{Compatibility with Active Learning}
\label{sec:al}

While \texttt{ADAPT} demonstrates significant improvements and can delay the onset of concept drift, at some point, labeled data may become necessary to handle drift effectively~\cite{kan2021investigating}. Active learning has proven to be one of the most effective approaches for addressing this issue in malware detection~\cite{chen2023continuous,jordaney2017transcend}. Although the combination of semi-supervised and active learning has been explored in low-data scenarios~\cite{tomanek2009semi,gao2020consistency}, its application to malware detection under concept drift remains largely unexplored. We explore this combination for the two Android malware detection datasets. 

We follow the approach from~\cite{tomanek2009semi}, where the samples with the highest uncertainty (i.e., the lowest predicted probability) are selected for human annotation using active learning, while high-quality pseudo-labels are selected using \texttt{ADAPT} at every retraining step. We incorporate active learning into \texttt{ADAPT} with a slight modification in Algorithm~\ref{alg:adapt}. We adopt the same active learning setup as in~\cite{chen2023continuous,alam2024revisiting}. Specifically, we first sample the $k$ most uncertain samples for human annotation for each test month, where $k$ represents the monthly annotation budget. These samples are then added to the labeled training dataset $D_l$, which increases in size by $k$ each month. Following this, we perform Step 1 in Algorithm~\ref{alg:adapt} to select pseudo-labeled samples using adaptive thresholding. However, we exclude the samples that have already been annotated through active learning, as we already have ground truth labels for them. We then continue with the rest of the algorithm and retrain the model. We perform a random hyperparameter search with 100 iterations, similar to~\cite{alam2024revisiting}.

Table~\ref{tab:res-al} presents the results with a monthly annotation budget of 50 on the Android datasets. For the XGBoost baseline model, we randomly select the same number of annotated samples per month to ensure consistency with the active learning methods following~\cite{apruzzese2022sok}.  We report active learning results based on uncertainty sampling, using XGBoost and the Hierarchical Contrastive Learning Classifier (HCC) introduced in~\cite{chen2023continuous}. HCC, the state-of-the-art method for active learning with neural networks, leverages contrastive learning with a hierarchical loss function to distinguish between samples from different malware families in the embedding space.

From Table~\ref{tab:res-al}, we observe that active learning improves performance over random selection for the XGBoost model. Furthermore, \texttt{ADAPT} enhances this performance by leveraging the remaining unlabeled data. Specifically, we observe a 0.4\% increase in F1-score for the Drebin dataset and a 0.7\% improvement for APIGraph. While these gains are relatively modest, they indicate that \texttt{ADAPT} can effectively benefit from active learning. 
Although our current approach involves a straightforward integration with active learning, exploring more sophisticated applications remains an avenue for future work.

\begin{table}[t]
\caption{Performance on multiclass classification (11-class) on BODMAS and EMBER datasets.}
    \centering
    \resizebox{\columnwidth}{!}{%
    \begin{tabular}{l|l|ccc}
        \toprule
        \textbf{Dataset} &
        {\textbf{Method}} & \textbf{F1} & \textbf{Precision} & \textbf{Recall} \\
        \midrule
        \multirow{5}{*}{BODMAS} & XGBoost & \underline{71.0±0.00} & 75.7±0.00 & 72.5±0.00 \\
        & ARF & 57.4±0.00 & 59.1±0.00 & 64.1±0.00 \\
        & MORSE & 68.1±2.21 & 67.8±4.68 & 73.1±0.78 \\
        & Insomnia & 70.1±1.21 & 71.6±0.99 & 73.6±1.70 \\
        & XGBoost+\texttt{ADAPT} & \textbf{\underline{71.4±0.65}} & 75.6±1.27 & 72.6±0.61 \\
        \midrule
        \multirow{5}{*}{EMBER} & XGBoost & \underline{62.4±0.00} & 70.6±0.00 & 68.7±0.00 \\
        & ARF & 11.0±0.00 & 11.8±0.00 & 17.7±0.00 \\
        & MORSE & 58.4±1.49 & 68.4±1.11 & 64.5±1.02 \\
        & Insomnia & 52.7±0.78 & 66.4±0.28 & 57.7±0.67 \\
        & XGBoost+\texttt{ADAPT} & \textbf{\underline{65.2±0.65}} & 70.3±0.94 & 71.4±0.59 \\
        \bottomrule
    \end{tabular}%
    }
    
    \label{tab:res-multiclass}
\end{table}


\begin{table}[t]
\caption{Performance on multiclass classification on BODMAS and EMBER datasets  with active learning.}
    \centering
    \resizebox{\columnwidth}{!}{%
    \begin{tabular}{l|l|ccc}
        \toprule
        \textbf{Dataset} &
        {\textbf{Method}} & \textbf{F1} & \textbf{Precision} & \textbf{Recall} \\
        \midrule
        \multirow{3}{*}{BODMAS} & XGBoost  & 73.1±0.37 & 77.6±0.29 & 74.7±0.61 \\
        & XGBoost + AL & 73.9±0.00 & 79.7±0.00 & 75.4±0.00 \\
        & XGBoost+\texttt{ADAPT} + AL & \textbf{\underline{76.1±0.68}} & 79.4±1.10 & 77.8±0.59 \\
        \midrule
        
        \multirow{3}{*}{EMBER} & XGBoost  & 78.0±0.51 & 80.6±0.35 & 82.4±0.54 \\
        & XGBoost + AL & 78.2±0.00 & 80.5±0.00 & 82.6±0.00 \\
        & XGBoost+\texttt{ADAPT} + AL & \textbf{\underline{80.6±1.00}} & 81.8±1.10 & 84.5±0.68 \\
        \bottomrule
    \end{tabular}%
    }
    
    \label{tab:res-multiclass-al-11}
\end{table}

\section{Extension to Multiclass Classification}
\label{sec:multiclass}

While we designed \texttt{ADAPT} primarily for binary malware detection, this section explores its applicability to a multiclass classification task. Specifically, we consider Windows malware family classification and construct two datasets from the BODMAS and EMBER datasets. 

\noindent\textbf{Datasets.}
We follow the same temporal split as in the binary classification task for training, validation, and test splits in both datasets. We select the top 10 most frequent malware families for each dataset in the training split and include the benign class, resulting in an 11-class classification task. We perform a hyperparameter search on the BODMAS dataset using 100 iterations per model and apply the selected hyperparameters directly to the EMBER dataset without further tuning. Appendix~\ref{app:multiclass} provides additional details on these datasets.



\noindent\textbf{Algorithm Adjustments.} 
In the binary detection setting, we use separate threshold parameters, $\tau_b$ for benign samples and $\tau_m$ for malware samples, updated dynamically based on model predictions on unlabeled data using the adaptation parameter $\lambda$. While $\lambda$ is shared across all classes, introducing separate thresholds for each malware family in the multiclass setting would significantly increase the number of parameters, complicating hyperparameter tuning. To mitigate this, we share $\tau_m$ across all malware classes, keeping the number of hyperparameters the same as in the binary setting (five), regardless of the number of malware families. Appendix~\ref{app:multiclass} explores a scenario without a benign class, where only malware families are classified. Apart from this threshold adjustment, the rest of the algorithm, including augmentation and mixup, remains unchanged.

\textbf{Results.}
We report the macro F1-score, precision, and recall in Table~\ref{tab:res-multiclass} for the XGBoost baseline, other self-training methods, and XGBoost+\texttt{ADAPT} on the two datasets. While \texttt{ADAPT} improves over the baseline on both datasets, other self-training methods perform worse than the XGBoost baseline. This effect is particularly pronounced in the EMBER dataset, which is more challenging and where we do not perform additional hyperparameter tuning. Notably, \texttt{ADAPT} improves the average F1-score over XGBoost by 2.8\%, whereas other self-training methods perform significantly worse than the baseline, suggesting that \texttt{ADAPT} is more robust to hyperparameter values compared to other methods on EMBER feature.

\textbf{Unknown Family Detection.}
To evaluate the open-set performance of the multi-class malware classifier, we conducted a monthly analysis with varying novelty in the unknown samples. For each month, we selected the top-$k$ most frequent unseen malware families ($k \in \{5, 10, 20\}$) and computed two key metrics: the \textit{evasion success rate}—the fraction of unknown samples misclassified as benign—and the \textit{AUC}, reflecting the model's ability to separate known from unknown samples based on confidence scores, in line with out-of-distribution (OOD) detection~\cite{yang2024generalized}. On the EMBER dataset, the XGBoost+\texttt{ADAPT} model achieved evasion success rates of 9.8\%, 7.1\%, and 6.6\% for 5, 10, and 20 new families, respectively, with corresponding AUC values of 0.706, 0.713, and 0.713. These results highlight a tendency to misclassify unknown malware as known malicious classes, though the classifier's AUC remains stable as novelty increases.


\textbf{Active Learning.}  
We conduct additional experiments using an annotation budget of 50 samples per month for active learning with the XGBoost model. Table~\ref{tab:res-multiclass-al-11} compares results for random sampling versus selecting the 50 most uncertain samples each month. Notably, integrating \texttt{ADAPT} with active learning improves performance over the random baseline by 3\% on the BODMAS dataset and 2.6\% on the EMBER dataset.


\section{Limitations \& Future Work}

\textbf{Evasion and Poisoning Attacks.} The effectiveness of our approach may be compromised by adversarial samples designed to evade malware detection classifiers~\cite{grosse2016adversarial,wang2023evasion}. While we do not explicitly address adversarial examples, integrating adversarial robustness techniques~\cite{bai2021recent} could enhance resilience. Additionally, poisoning attacks, where an adversary injects carefully crafted samples into the unlabeled data pool, can impact continual learning updates under concept drift~\cite{taheri2020defending,korycki2023adversarial}. Such poisoning can cause the pseudo-labeling process to reinforce incorrect predictions, increasing the risk of persistent false negatives or false positives. Although semi-supervised learning methods may offer some resistance to random label noise~\cite{wu2023grim}, defending against targeted adversarial poisoning remains an open challenge~\cite{goldblum2022dataset}. We leave the exploration of adversarial robustness and poisoning-aware defenses in pseudo-labeling pipelines as important directions for future work.

\textbf{Computational Efficiency.}
\texttt{ADAPT} introduces no additional latency during deployment, as its inference time remains identical to that of the baseline model. However, the training time increases linearly due to data augmentation and mixup strategies. We provide a detailed analysis of the computational overhead in Appendix~\ref{app:compute}.

\textbf{Limitations in Features.}
Our experiments rely on static analysis features, which may not be sufficient for effectively handling concept drift~\cite{dambra2023decoding}. Incorporating dynamic features, or a combination of static and dynamic features, could potentially improve the model's ability to adapt to concept drift. 
Its effectiveness with dynamic or hybrid features can be explored in future studies. 


\textbf{Catastrophic forgetting.} Catastrophic forgetting refers to the tendency of a learned model to lose previously acquired knowledge when trained on new data—a challenge particularly relevant in malware detection~\cite{rahman2022limitations}. We analyze its impact on the Android malware detection datasets in Appendix~\ref{app:cf}. While our approach does not directly address catastrophic forgetting, future work could jointly integrate mechanisms to handle concept drift and mitigate forgetting. One promising direction is using replay-based methods from continual learning~\cite{rolnick2019experience}. In our context, this could involve maintaining a replay buffer of class-balanced, confidently pseudo-labeled samples from intermediate time steps, which can be reused during subsequent model fine-tuning.

\textbf{Applicability to Other Security Domains.}  
Future work could explore extending \texttt{ADAPT} to other security domains, such as intrusion and phishing detection, where concept drift occurs frequently. While effective data augmentation may require domain-specific customization, other key components of \texttt{ADAPT}, such as adaptive thresholding and mixup regularization, remain domain-independent.

\section{Conclusion}
\label{sec:conc}
We present \texttt{ADAPT}, a pseudo-labeling-based approach to mitigate concept drift in malware detection. Our model-agnostic framework is validated across five diverse datasets exhibiting different types and degrees of distributional shift. Through comprehensive experiments, we show that \texttt{ADAPT} overcomes limitations of prior pseudo-labeling strategies and consistently improves detection performance. Our method also integrates well with active learning, further enhancing adaptability in practice. Overall, \texttt{ADAPT} provides a practical, label-efficient, and robust solution for maintaining detection accuracy under evolving threats, and its principles are readily extensible to other security and drift-sensitive tasks.

\section*{Acknowledgment}
The authors acknowledge Research Computing at the Rochester Institute of Technology~\cite{rit2025researchcomputing} for providing computational resources and support that have contributed to the research results reported in this publication.

\clearpage

\bibliographystyle{plain}
\bibliography{references}

\clearpage

\appendices


\section{Self-Training Baselines}
\label{app:st}

DroidEvolver~\cite{xu2019droidevolver} and its improved version, DroidEvolver++ (DE++)~\cite{kan2021investigating}, are notable self-training approaches for Android malware detection, but both suffer from self-poisoning and limited evaluation across datasets. Beyond these, research on self-training for concept drift in malware detection remains scarce, motivating our inclusion of additional established self-training algorithms for comparison.

\begin{enumerate}
    \item \textbf{Adaptive Random Forest (ARF)~\cite{gomes2017adaptive}}: ARF is an online ensemble method for concept drift adaptation. To enable self-training, we introduce a threshold parameter for pseudo-label selection, which is tuned in conjunction with other tree hyperparameters, allowing ARF to update using pseudo-labeled samples without requiring ground-truth labels.

    \item \textbf{DroidEvolver++ (DE++)~\cite{kan2021investigating}}: DE++ improves upon DroidEvolver~\cite{xu2019droidevolver} for malware drift adaptation without requiring ongoing ground truth. It uses an ensemble of five linear models to generate pseudo-labels and tracks model aging with a buffer of recent samples. When prediction consistency in the buffer drops below a threshold, the model is updated using new pseudo-labels from the ensemble.

    \item \textbf{Insomnia~\cite{andresini2021insomnia}}: Insomnia adapts to concept drift in network intrusion detection via co-training between a Nearest Centroid (NC) classifier and an MLP, with the NC providing pseudo-labels for uncertain MLP predictions. Both models are updated using pseudo-labeled data. For malware detection, we found NC to be ineffective, so we use XGBoost with the MLP, and select the most confident samples for updating to avoid early self-poisoning.

    \item \textbf{MORSE}~\cite{wu2023grim}: MORSE is a state-of-the-art semi-supervised method for malware family classification with noisy labels. It uses the FixMatch~\cite{sohn2020fixmatch} framework, incorporating an augmentation strategy that enforces consistency between weakly and strongly augmented versions of each pseudo-labeled sample. We include MORSE as a baseline due to its strong performance in the presence of label noise.

\end{enumerate}


The first two baselines, ARF and DE++, are online methods that update one sample at a time and do not require access to the full training dataset after initialization. In contrast, Insomnia and MORSE use batch training, updating models by combining original training data with pseudo-labeled samples. While our main experiments assume the training data remains available, we also consider restricted scenarios without access to the original training data (see Appendix~\ref{sec:sf-adapt}).

\section{Multiclass Classification}
\label{app:multiclass}

\begin{table}[t]
\caption{Statistics of the BODMAS multiclass classification dataset}
    \centering
    \begin{tabular}{ccccc}
        \toprule
        \textbf{ID} & \textbf{Family} & \makecell{\textbf{Train} \\ \textbf{Samples}} & \makecell{\textbf{Validation} \\ \textbf{Samples}} & \makecell{\textbf{Test} \\ \textbf{Samples}} \\
        \midrule
        0  & Benign     & 3000  & 3000  & 6000  \\
        1  & Wacatac    & 2337  & 917   & 988   \\
        2  & Upatre     & 930   & 1148  & 1369  \\
        3  & Mira       & 586   & 512   & 791   \\
        4  & Small      & 558   & 1026  & 1475  \\
        5  & Dinwod     & 405   & 682   & 701   \\
        6  & Wabot      & 368   & 748   & 2405  \\
        7  & Autorun    & 363   & 148   & 60    \\
        8  & Musecador  & 309   & 717   & 19    \\
        9  & Gepys      & 298   & 382   & 255   \\
        10 & Berbew     & 284   & 195   & 1195  \\
        \bottomrule
    \end{tabular}
        
    \label{tab:data_bodmas_fam}
\end{table}

\begin{table}[t]
\caption{Statistics of the EMBER multiclass classification dataset}
    \centering
    \begin{tabular}{cccc}  
        \toprule
        \textbf{ID} & \textbf{Family} & \makecell{\textbf{Train} \\ \textbf{Samples}} & \makecell{\textbf{Test} \\ \textbf{Samples}} \\
        \midrule
        0  & Benign            & 500  & 5500 \\
        1  & InstallMonster    & 500  & 3217 \\
        2  & AdPoshel          & 500  & 2647 \\
        3  & Zusy              & 500  & 3783 \\
        4  & Fareit            & 500  & 4662 \\
        5  & Emotet            & 500  & 5386 \\
        6  & DealPly           & 500  & 2227 \\
        7  & DotDo             & 500  & 953  \\
        8  & StartSurf         & 500  & 3819 \\
        9  & Mira              & 425  & 304  \\
        10 & Tiggre            & 405  & 1284 \\
        \bottomrule
    \end{tabular}

    \label{tab:stat_ember_fam}
\end{table}

\subsection{Dataset Statistics}

Table~\ref{tab:data_bodmas_fam} and Table~\ref{tab:stat_ember_fam} present class-wise statistics for the BODMAS and EMBER datasets, respectively.

For BODMAS, we utilize all available samples per malware family in training and randomly sample 1,000 benign samples per month across all splits. Some families, like Gepys, are rare in the test set, while others, such as Berbew, are more frequent during testing than training. For EMBER, we cap each class, including Benign, at 500 samples in training and per month in testing to manage dataset size. This produces approximately balanced training data but class imbalance in the test set.

\subsection{Experiments Excluding Benign Class}
In practice, malware analysis typically proceeds in two steps: first, binary classification distinguishes between benign and malicious samples; second, if a sample is identified as malware, its family is determined~\cite{rahman2022limitations}. We emulate this scenario for malware family classification by removing benign samples from the BODMAS (Table~\ref{tab:data_bodmas_fam}) and EMBER (Table~\ref{tab:stat_ember_fam}) datasets, resulting in a 10-class task. In this setting, \texttt{ADAPT} uses a single threshold parameter $\tau_m$ shared across all malware families; otherwise, the algorithm is unchanged from the binary case. This reduces the number of \texttt{ADAPT} hyperparameters from five to four.

\begin{table}[t]
\caption{Performance on multiclass classification (10-class) on BODMAS and EMBER datasets.}
    \centering
    \resizebox{\columnwidth}{!}{%
    \begin{tabular}{l|l|ccc}
        \toprule
        \textbf{Dataset} &
        {\textbf{Method}} & \textbf{F1} & \textbf{Precision} & \textbf{Recall} \\
        \midrule
        \multirow{5}{*}{BODMAS} & XGBoost & \underline{76.8±0.00} & 82.9±0.00 & 78.3±0.00 \\
        & ARF & 63.5±0.00 & 70.6±0.00 & 64.0±0.00 \\
        & MORSE & 75.3±0.86 & 80.1±1.18 & 76.4±1.09 \\
        & Insomnia & 62.2±1.19 & 69.5±1.47 & 65.7±1.44 \\
        & XGBoost+\texttt{ADAPT} & \textbf{\underline{77.0±1.07}} & 82.2±1.26 & 77.7±0.99 \\
        \midrule
        \multirow{5}{*}{EMBER} & XGBoost & \underline{58.7±0.00} & 67.8±0.00 & 66.3±0.00 \\
        & ARF & 10.9±0.00 & 9.66±0.00 & 21.1±0.00 \\
        & MORSE & 53.6±2.23 & 62.8±1.68 & 63.1±0.89 \\
        & Insomnia & 39.6±0.53 & 58.5±1.31 & 49.0±0.23 \\
        & XGBoost+\texttt{ADAPT} & \textbf{\underline{65.2±1.43}} & 75.1±1.94 & 70.4±1.35 \\
        \bottomrule
    \end{tabular}%
    }
    
    \label{tab:res-multiclass-10}
\end{table}

\begin{table}[t]
\caption{Performance on multiclass classification (10-class) on BODMAS and EMBER datasets  with active learning.}
    \centering
    \resizebox{\columnwidth}{!}{%
    \begin{tabular}{l|l|ccc}
        \toprule
        \textbf{Dataset} &
        {\textbf{Method}} & \textbf{F1} & \textbf{Precision} & \textbf{Recall} \\
        \midrule
        \multirow{3}{*}{BODMAS} & XGBoost  & 81.2±0.96 & 86.2±0.99 & 81.9±0.96 \\
        & XGBoost + AL & 83.2±0.00 & 89.2±0.00 & 83.6±0.00 \\
        & XGBoost+\texttt{ADAPT} + AL & \textbf{\underline{85.5±1.08}} & 90.6±0.37 & 85.8±1.58 \\
        \midrule
        
        \multirow{3}{*}{EMBER} & XGBoost  & 76.4±1.22 & 79.2±1.06 & 80.6±1.21 \\
        & XGBoost + AL & 77.1±0.00 & 79.6±0.00 & 81.8±0.00 \\
        & XGBoost+\texttt{ADAPT} + AL & \textbf{\underline{83.8±0.65}} & 85.0±0.55 & 87.2±0.40 \\
        \bottomrule
    \end{tabular}%
    }
    
    \label{tab:res-multiclass-al}
\end{table}

Table \ref{tab:res-multiclass-10} presents the results of \texttt{ADAPT} and other self-training approaches across both datasets. \texttt{ADAPT} outperforms the baseline XGBoost methods on both datasets, while other self-training methods perform worse. The improvement is especially notable on the EMBER dataset, where \texttt{ADAPT} achieves a 6.5\% increase in the F1-score.
Table~\ref{tab:res-multiclass-al} presents the results for active learning with a monthly annotation budget of 50 samples. Active learning with uncertainty sampling outperforms the baseline XGBoost model trained with the same annotation budget using random sampling. Additionally, incorporating \texttt{ADAPT} further enhances performance on both datasets. Specifically, integrating \texttt{ADAPT} improves the F1-score by 2.3\% on the BODMAS dataset and 6.1\% on the EMBER dataset, demonstrating the effectiveness of leveraging unlabeled samples even in the presence of active learning.





\section{Support Vector Machine(SVM) Experiments}  

We conduct additional experiments using SVM baseline models on two Android datasets, APIGraph and Drebin, using the same hyperparameter selection strategy from Section~\ref{sec:hyperparameters}. The results, including those with \texttt{ADAPT}, are summarized in Table~\ref{tab:svm-res}.  

The baseline SVM model performs competitively with the baseline XGBoost model on the APIGraph dataset. However, its performance is significantly lower on the Drebin dataset than the XGBoost baseline.  

Integrating \texttt{ADAPT} with the SVM model proves to be effective, yielding performance improvements over the baseline SVM across both datasets. Notably, \texttt{ADAPT} significantly reduces the false negative rate,  improving the F1-score. This effect is particularly pronounced in the Drebin dataset, where we observe an average F1-score increase of 14.2\%.

\begin{table}[t]
\caption{SVM performance on APIGraph and Drebin (average over five runs).}
\centering
\begin{tabular}{lcccccc}
\toprule
 & \multicolumn{3}{c}{\textbf{APIGraph}} & \multicolumn{3}{c}{\textbf{Drebin}} \\
\cmidrule(lr){2-4} \cmidrule(lr){5-7}
\textbf{Method} & F1 & FPR & FNR & F1 & FPR & FNR \\
\midrule
SVM & 66.3 & 1.41 & 39.5 & 33.8 & 0.70 & 76.7 \\
SVM+ADAPT & 73.6 & 1.47 & 29.3 & 48.0 & 0.84 & 61.5 \\
\bottomrule
\end{tabular}
\label{tab:svm-res}
\end{table}

\section{Source-Free Adaptation}
\label{sec:sf-adapt}

\begin{figure}[t]
    \centering
    \includegraphics[width=\linewidth]{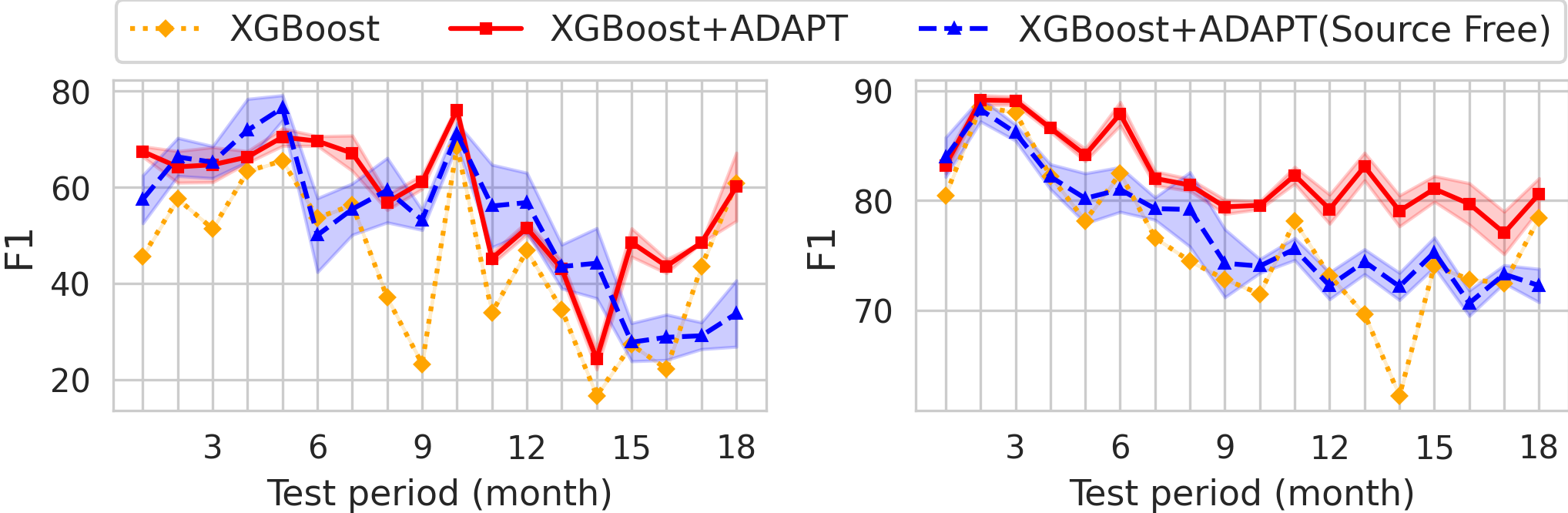}
    \caption{F1 score on Drebin (left) and APIGraph (right) datasets across the first 18 test months for XGBoost, XGBoost with \texttt{ADAPT}, and XGBoost with \texttt{ADAPT} in a source-free setting (i.e., without access to the training data during retraining).}
    \label{fig:source-free}
\end{figure}


In some scenarios, access to the original training data may be unavailable after initial model training due to privacy, proprietary, storage, or computational constraints~\cite{xu2019droidevolver}. In these cases, model updates must rely solely on pseudo-labeled data. We explore this setting for XGBoost on the Drebin and APIGraph datasets, modifying Algorithm~\ref{alg:adapt} at line 7 so that \( D_m = D_p \), meaning only pseudo-labeled data are available after the initial phase. Separate hyperparameter optimization is performed for this restricted scenario using the same search space.

Figure~\ref{fig:source-free} compares the baseline XGBoost model, the ADAPT semi-supervised update, and source-free adaptation on these datasets. For APIGraph, results are shown for the first 18 test months, as performance declines over longer periods (average F1: 61.24\%), though the source-free method offers noticeable gains in this window. On Drebin, source-free adaptation improves the average F1 by 11.3\% over the baseline in the first year, lagging the semi-supervised variant by just 1.76\%. On APIGraph, it outperforms the baseline by 0.82\% but is 3.97\% below the semi-supervised approach. These findings indicate that \texttt{ADAPT} can still mitigate concept drift when the original training data is unavailable, making it suitable for continual learning scenarios with storage limitations~\cite{rahman2022limitations}.



\begin{figure}[t]

    \centering

    \begin{subfigure}[t]{0.99\linewidth}
        \centering
        \includegraphics[width=\linewidth]{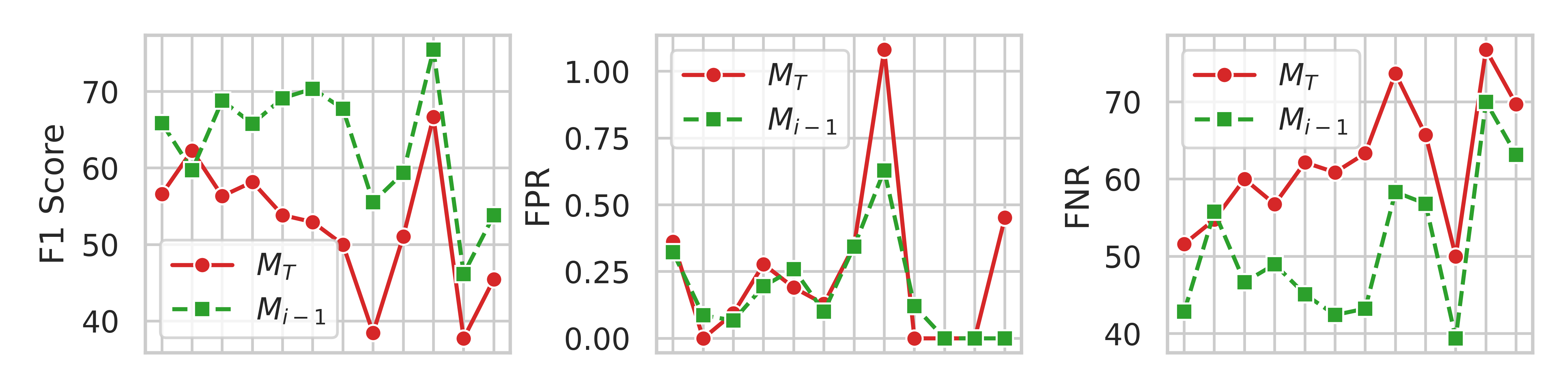}
        \caption{XGBoost}
    \end{subfigure}
    
    \vspace{-0.05cm} 
    \begin{subfigure}[t]{0.99\linewidth}
        \centering
        \includegraphics[width=\linewidth]{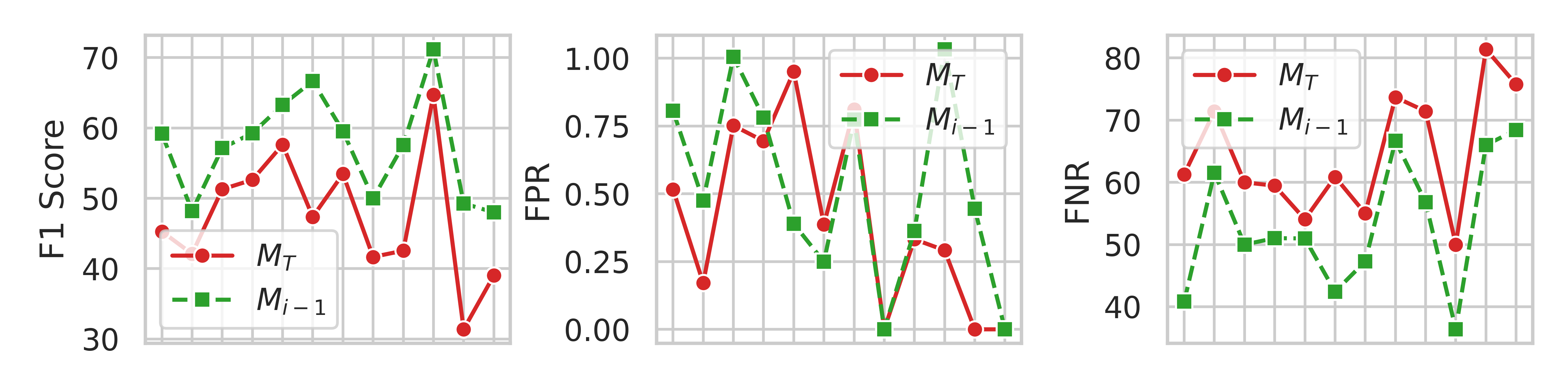}
        \caption{MLP}
    \end{subfigure}

    \caption{Impact of catastrophic forgetting on Drebin: XGBoost and MLP performance over the first 12 test months. $M_T$ is the final-month model; $M_{i-1}$ is the previous month’s model.}

    \label{fig:cf-drebin}
\end{figure}

\begin{figure}[t]
    \centering

    \begin{subfigure}[t]{0.99\linewidth}
        \centering
        \includegraphics[width=\linewidth]{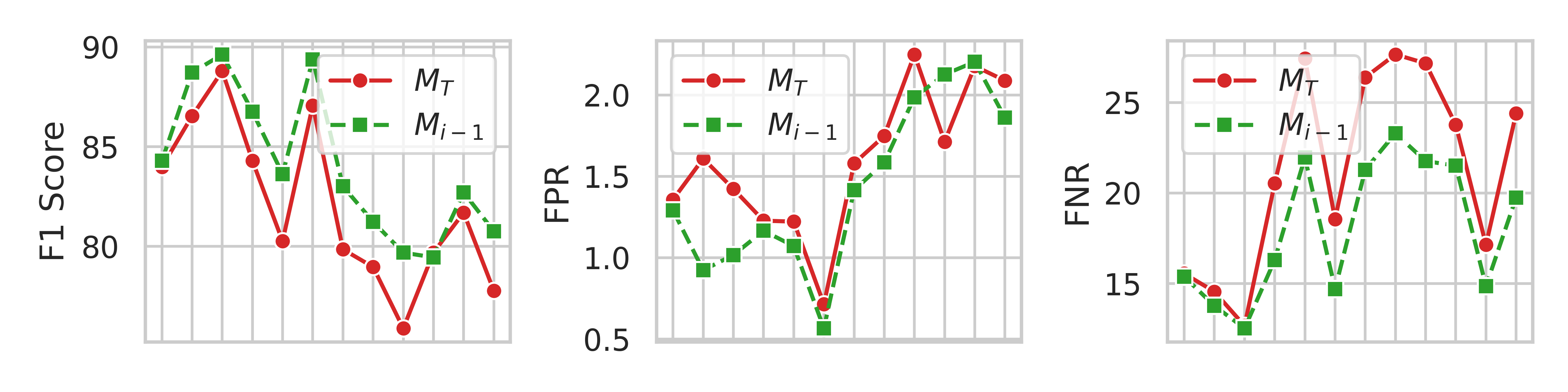}
        \caption{XGBoost}
    \end{subfigure}
    
    \vspace{-0.05cm} 
    \begin{subfigure}[t]{0.99\linewidth}
        \centering
        \includegraphics[width=\linewidth]{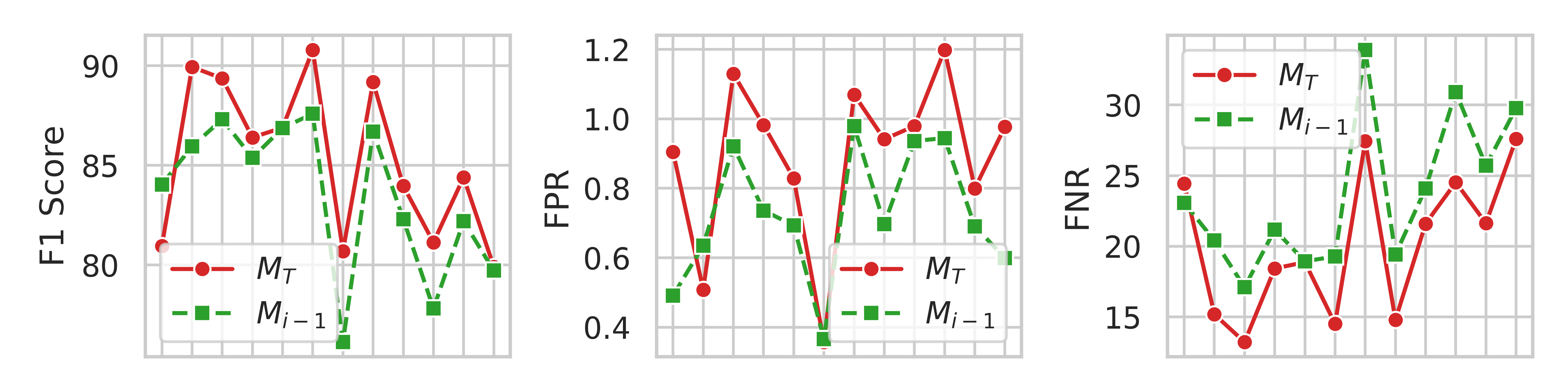}
        \caption{MLP}
    \end{subfigure}

    \caption{Impact of catastrophic forgetting on the APIGraph dataset for XGBoost and MLP models.}
    \label{fig:cf-apigraph}
\end{figure}

\section{Catastrophic Forgetting}
\label{app:cf}

Catastrophic forgetting, where a model loses previously acquired knowledge when learning from new data, is a significant challenge in continual learning~\cite{rahman2022limitations}. While our work does not directly address this issue, we assess its impact on the Android malware datasets using XGBoost and MLP. Specifically, we evaluate the final model trained on the last month (\(M_T\)) against the first year of test data, comparing its F1 score, FPR, and FNR to those of models trained during each monthly retraining phase.

The results are presented in Figure~\ref{fig:cf-drebin} for the Drebin dataset and Figure~\ref{fig:cf-apigraph} for the APIGraph dataset. Both models exhibit some degree of catastrophic forgetting on the Drebin dataset, as the final model’s performance degrades compared to earlier models. In contrast, on the APIGraph dataset, the final MLP model outperforms its earlier versions, indicating no signs of catastrophic forgetting. For the XGBoost model, catastrophic forgetting is primarily driven by an increase in false negatives, whereas for the MLP model, the FNR increases on Drebin but decreases on APIGraph.

\section{Computational Efficiency}
\label{app:compute}

While \texttt{ADAPT} does not increase inference time, it incurs additional training cost, which increases approximately linearly due to data augmentation and mixup. This overhead is primarily driven by two factors: (1) the time required to generate augmented samples and (2) the additional time needed to train on the expanded dataset, roughly three times larger than the original training set.

The extent of this increase depends on the dataset size, feature dimensionality, and model complexity. Table~\ref{tab:compute} reports the training time overhead introduced by \texttt{ADAPT} across different models and datasets. The most significant increase is observed for the XGBoost model on the Drebin dataset, where the high feature dimensionality (16{,}978) substantially impacts computation time. For other datasets, training time typically increases by three to five relative to the baseline model.

\begin{table}[t]
\caption{Training time (in seconds) of different baseline models with \texttt{ADAPT}}
    \centering
    \resizebox{0.99\columnwidth}{!}{%

    \begin{tabular}{l|cc|cc|cc}
        \toprule
        Dataset & RF & \makecell{RF +\\  \texttt{ADAPT}} & MLP & \makecell{MLP +\\  \texttt{ADAPT}} & XGBoost & \makecell{XGBoost +  \\  \texttt{ADAPT}} \\
        \midrule
        Drebin   & 38.5  & 281.0  & 158.0  & 712.9  & 90.3  & 1378.6  \\
        APIGraph & 19.5  & 105.0  & 32.0   & 37.0   & 6.2   & 17.2    \\
        BODMAS   & 45.9  & 188.3  & 242.9  & 681.6  & 93.5  & 301.6   \\
        EMBER    & 111.7 & 422.1  & 512.7  & 1618.5 & 237.4 & 668.1   \\
        PDF      & 12.5  & 57.5   & 38.1   & 70.4   & 6.1   & 66.0    \\
        \bottomrule
    \end{tabular}
    }

    \label{tab:compute}
\end{table}

\section{Hyperparameter Search}
\label{sec-app--hpo}

Table~\ref{tab:hyperparameters} summarizes the hyperparameter ranges used for each method. For self-training baselines, we adopt default parameters for the five online learning classifiers. Insomnia, MORSE, and our method \texttt{ADAPT} use different underlying models: Insomnia applies co-training with XGBoost and MLP, MORSE uses MLP, and \texttt{ADAPT} is evaluated with Random Forest, XGBoost, and MLP. All models share the same parameter search space as in the offline setting. We perform a random search with 200 trials across the joint parameter space to select the best hyperparameters for each dataset.

\begin{table*}[t]
\caption{Hyperparameter search spaces for various methods. "U" indicates sampling from a uniform random distribution within the specified range. Hyperparameters marked with "*" are only used during the retraining phase. The Insomnia method also uses hyperparameters from MLP and XGBoost, while MORSE uses those from MLP. \texttt{ADAPT} utilizes the baseline model, which can be either Random Forest (RF), XGBoost, or MLP.}
\centering
\resizebox{\textwidth}{!}{  
\begin{tabular}{p{3cm} p{7cm} p{7cm}} 
\toprule
\textbf{Model} & \textbf{Hyperparameter} & \textbf{Candidate Values} \\
\midrule
\multirow{4}{*}{Random Forest} & n\_estimators &  $2^x$, where $x \in \text{U}[5, 10]$ \\
 & max\_depth & $2^y$, where $y \in \text{U}[5, 10]$ \\
 & criterion &  \{gini, entropy, log\_loss\} \\
 & class\_weight & \{None, "balanced"\} \\
\midrule
\multirow{6}{*}{XGBoost} & max\_depth & $2^w$, where $w \in \text{U}[3, 7]$ \\
 & alpha & $10^a$, where $a \in \text{U}[-8, 0]$ \\
 & lambda & $10^b$, where $b \in \text{U}[-8, 0]$ \\
 & eta & $3.0 \times 10^c$, where $c \in \text{U}[-2, -1]$ \\
 & balance & \{True, False\} \\
 & num\_boost\_round & \{100, 150, 200, 300, 400\} \\
\midrule
\multirow{7}{*}{MLP} & mlp\_layers & \{[100, 100], [512, 256, 128], [512, 384, 256, 128], [512, 384, 256, 128, 64]\} \\
 & learning\_rate & $10^d$, where $d \in \text{U}[-5, -3]$ \\
 & dropout & $x$, where $x \in \text{U}[0.0, 0.5]$ \\
 & batch\_size & $2^e$, where $e \in $ \{5, 6, 7, 8, 9, 10\} \\
 & epochs & \{25, 30, 35, 40, 50, 60, 80, 100, 150\} \\
 & optimizer & \{Adam\} \\
 & balance & \{True, False\} \\
 & cont\_learning\_epochs* & \{0.1, 0.2, 0.3, 0.4, 0.5\} \\
\midrule
\multirow{2}{*}{SVM} & C & $10^z$, where $z \in \text{U}[-4, 3]$ \\
 & class\_weight & \{None, "balanced"\} \\
 \midrule
\multirow{5}{*}{ARF} & n\_models & $2^x$, where $x \in \text{U}[3, 5]$ \\
 & max\_features & \{sqrt, log2, None\} \\
 & max\_depth & $2^y$, where $y \in \text{U}[5, 10]$ \\
 & lambda\_value & 6 \\
 & threshold & $x$, where $x \in \text{U}[0.6, 0.99]$ \\
\midrule
\multirow{3}{*}{DE++} & age\_threshold\_low & $x$, where $x \in \text{U}[0.0, 0.5]$ \\
 & buffer\_ratio & $x$, where $x \in \text{U}[0.1, 0.5]$ \\
 & buffer\_size & \{1000, 2000, 3000\} \\
\midrule
\multirow{1}{*}{Insomnia} & fraction & $x$, where $x \in \text{U}[0.0, 0.8]$ \\
\midrule
\multirow{3}{*}{MORSE} & threshold & $x$, where $x \in \text{U}[0.8, 0.99]$ \\
 & weak\_augment & $x$, where $x \in \text{U}[0.0, 0.1]$ \\
 & strong\_augment & $x$, where $x \in \text{U}[0.1, 0.2]$ \\
\midrule
\multirow{5}{*}{\texttt{ADAPT}} & threshold\_benign & $x$, where $x \in \text{U}[0.8, 0.99]$ \\
 & threshold\_malware & $x$, where $x \in \text{U}[0.6, 0.99]$ \\
 & lambda & $x$, where $x \in \text{U}[0.0, 0.5]$ \\
 & mask\_ratio & $x$, where $x \in \text{U}[0.0, 0.2]$ \\
 & mixup\_alpha & $x$, where $x \in \text{U}[0.0, 0.2]$ \\
\bottomrule
\end{tabular}
}

\label{tab:hyperparameters}
\end{table*}

\section{Theoretical Analysis}
\label{sec:theory}
Our self-training approach for malware detection under concept drift is theoretically supported by prior work on gradual self-training~\cite{kumar2020understanding}. We outline key theoretical results explaining its effectiveness under distribution shifts.

\subsection{Setup and Assumptions}  
We consider a gradual domain shift scenario, characterized by a sequence of distributions $P_0, P_1, \dots, P_T$, where $P_0$ is the source domain and $P_T$ the target. The shift is assumed gradual: for a small $\epsilon > 0$, the distance $\rho(P_t, P_{t+1}) < \epsilon$ for all $0 \leq t < T$, with $\rho$ denoting a distributional distance metric. We assume there exists a sequence of classifiers $\theta_t$ that can correctly classify most samples with a margin, and that the incremental shifts between domains permit effective iterative adaptation.

\subsection{Main Theoretical Results}  
The primary result from~\cite{kumar2020understanding} states that under gradual domain shift, self-training can iteratively reduce error. Specifically:

\begin{theorem}[\cite{kumar2020understanding}, Theorem 3.2]

Let $P$ and $Q$ be two distributions such that the Wasserstein distance~\cite{villani2008optimal} satisfies $\rho(P, Q)=\rho<1/R$, where $R$ is the regularization parameter. Assume that both distributions have the same marginal distribution over labels, i.e., $P(Y) = Q(Y)$. 

Suppose we have an initial model $\theta \in \Theta_R$ trained on $P$, and we perform self-training using $n$ unlabeled samples from $Q$ to obtain an updated model $\theta'$. Then, with probability at least $1 - \delta$ over the sampling of $n$ unlabeled examples from $Q$, the ramp loss~\cite{brooks2011support} of the new model $\theta'$ on $Q$ satisfies:

\begin{align}
L_r(\theta', Q) &\leq \frac{2}{1 - \rho R} L_r(\theta, P) + \alpha^* \notag \\
&\quad + \frac{4BR + \sqrt{2 \log (2/\delta)}}{\sqrt{n}}, \label{eq:self-training-bound}
\end{align}

Here, $L_r(\theta, P)$ represents the ramp loss of the initial model on the source distribution $P$, while $\alpha^* = \min_{\theta^* \in \Theta_R} L_r(\theta^*, Q)$ denotes the minimum achievable ramp loss on the target distribution $Q$ within the hypothesis space $\Theta_R$. The term $B$ is an upper bound on the norm of the input features, ensuring bounded feature magnitudes. The parameter $R$ controls the capacity of the hypothesis space by regularizing the complexity of the learned models. Finally, $\delta$ is the confidence parameter that determines the probability with which the bound holds over the sampling of $n$ unlabeled examples from $Q$.

\end{theorem}

This theorem implies that if the initial model has low loss on the source domain and the distributional shift is gradual, self-training will yield a model with controlled error on the new distribution. Applying this argument iteratively across $T$ steps leads to an exponential improvement over direct adaptation:

\begin{corollary}[\cite{kumar2020understanding}, Corollary 3.3]
\label{cor:gradual-self-training}
Under the assumptions of $\alpha^*$-separation, no label shift, gradual shift, and bounded data, suppose the initial model $\theta_0$ has a ramp loss of at most $\alpha_0 \geq \alpha^*$ on the source distribution $P_0$, i.e., $L_r(\theta_0, P_0) \leq \alpha_0$. If self-training is applied iteratively over $T$ steps to produce the final model $\theta_T$, denoted as $\theta_T = ST(\theta_0, (S_1, \dots, S_T))$, then with high probability, the ramp loss on the final target distribution $P_T$ is bounded as:

\begin{align}
L_r(\theta_T, P_T) &\leq \beta^{T+1} \Bigg( \alpha_0 +  
\frac{4BR + \sqrt{2 \log (2T / \delta)}}{\sqrt{n}} \Bigg) \notag \\
&\quad \label{eq:gradual-self-training-bound}
\end{align}

where $\beta = \frac{2}{1 - \rho R}$. This result demonstrates that gradual self-training effectively controls the error over multiple adaptation steps, preventing the potential failure cases that can arise from direct adaptation.

\end{corollary}

\subsection{Extension to ADAPT}

In the malware classification setting, we extend the theoretical framework by considering different distribution shifts for the two classes: malware and benign. Let $\rho_m$ and $\rho_b$ denote the Wasserstein shifts for the malware and benign classes, respectively, where we assume $\rho_m > \rho_b$. This reflects the practical scenario where malware samples undergo more significant distributional changes compared to benign samples.





\subsubsection{Asymmetric Error}
Following Lemma A.2 from~\cite{kumar2020understanding}, we analyze class-specific error bounds under different shift magnitudes. The classification error on the target distribution $Q$ for malware and benign classes is bounded by:
\begin{align}
    \text{Err}^m(\theta, Q) &\leq \frac{1}{1 - \rho_m R} L_r^m(\theta, P), \notag \\
    \text{Err}^b(\theta, Q) &\leq \frac{1}{1 - \rho_b R} L_r^b(\theta, P). \label{eq:class-specific-error}
\end{align}
Since $\rho_m > \rho_b$, the multiplicative factor for malware is larger, meaning that—even with similar initial losses—classification error for malware will be higher than for benign samples under shift.

Consequently, malware samples experience greater drift and are more likely to be misclassified as benign, increasing the false negative rate (FNR). By contrast, benign samples cross into the malware region less frequently, resulting in a lower false positive rate (FPR). This asymmetry explains why FNR typically exceeds FPR in malware classification under distributional shift.

\subsubsection{Impact of Class-Specific Thresholding}

To mitigate the imbalance between false negative and false positive rates, we introduce separate thresholds $\tau_m$ and $\tau_b$ for malware and benign classifications, respectively. Instead of classifying all samples, we only classify those for which the model's confidence exceeds the respective threshold. This effectively reduces the number of classified samples but also lowers the classification error.

Formally, let $Q_{\tau_m}^m$ and $Q_{\tau_b}^b$ denote the distributions of malware and benign samples that remain after thresholding. Since thresholding removes uncertain samples near the decision boundary, the ramp loss on the remaining classified samples decreases:

\begin{align}
    L_r^m(\theta, Q_{\tau_m}^m) \leq L_r^m(\theta, Q), \notag \\
    L_r^b(\theta, Q_{\tau_b}^b) \leq L_r^b(\theta, Q). \label{eq:threshold-loss}
\end{align}

Applying the error bound from Equation~\eqref{eq:class-specific-error} to the thresholded samples, we obtain:

\begin{align}
    \text{Err}^m(\theta, Q) &\leq \frac{1}{1 - \rho_m R} L_r^m(\theta, P_{\tau_m}), \notag \\
    \text{Err}^b(\theta, Q) &\leq \frac{1}{1 - \rho_b R} L_r^b(\theta, P_{\tau_b}). \label{eq:threshold-error}
\end{align}

Since thresholding removes samples that are close to the decision boundary—where errors are more likely—the filtered distributions $Q_{\tau_m}^m$ and $Q_{\tau_b}^b$ contain fewer misclassified samples, leading to lower classification errors. 

Thresholding reduces the number of classified samples, which affects the generalization bound in Equation~\eqref{eq:self-training-bound}. Specifically, the third term in the bound, which depends on the sample size $n$, is modified as follows:

\begin{align}
\frac{4BR + \sqrt{2 \log (2/\delta)}}{\sqrt{n}} \quad \rightarrow \quad \frac{4BR + \sqrt{2 \log (2/\delta)}}{\sqrt{n_{\tau_m, \tau_b}}}, \label{eq:threshold-effect}
\end{align}

where $n_{\tau_m, \tau_b}$ is the number of samples retained after thresholding at $\tau_m$ and $\tau_b$, and we have $n_{\tau_m, \tau_b} < n$. Since the denominator decreases, this term increases, leading to a looser bound.

\subsubsection{Modified Error Bound}

Following Lemma A.3 from~\cite{kumar2020understanding}, we account for the fact that the total error depends on the mixture of malware and benign samples in \( Q \). Given that \( q_m \) and \( q_b \) represent the proportions of malware and benign samples in \( Q \) (i.e., \( q_m + q_b = 1 \)), we replace the expectation in the bound with an explicit weighted sum.  

First, the error term in Equation~\eqref{eq:self-training-bound} becomes:  

\begin{align}
\frac{2}{1 - \rho R} L_r(\theta, P) 
\quad &\rightarrow \quad  
q_m \frac{2}{1 - \rho_m R} L_r^m(\theta, P_{\tau_m}) \notag \\
&\quad + q_b \frac{2}{1 - \rho_b R} L_r^b(\theta, P_{\tau_b}).  
\label{eq:expected-error-term}
\end{align}

Thus, the final modified bound becomes:  

\begin{align}
L_r(\theta', Q_{\tau_m, \tau_b}) &\leq  
q_m \frac{2}{1 - \rho_m R} L_r^m(\theta, P_{\tau_m}) \notag \\
&\quad + q_b \frac{2}{1 - \rho_b R} L_r^b(\theta, P_{\tau_b}) \notag \\
&\quad + \alpha^* \notag \\
&\quad + \frac{4BR + \sqrt{2 \log (2/\delta)}}{\sqrt{n_{\tau_m, \tau_b}}}.  
\label{eq:the-final}
\end{align}



\subsubsection{Impact of Augmentation \& Mixup}
Data augmentation and mixup serve as effective regularizers during training~\cite{lin2024good,zhang2020does}, helping to reduce overfitting and improve generalization. In our theoretical framework, this regularization implicitly reduces the norm of the learned weight vector $w$, leading to a smaller capacity parameter $R$ (i.e., $\|w\|_2 \leq R$). A lower $R$ constrains model complexity, making it less likely to fit noise or learn overly intricate decision boundaries. Consequently, stronger regularization tightens the bound in Equation~\ref{eq:the-final} and enhances robustness to distributional shifts. A detailed theoretical analysis of these effects is left for future work.

\end{document}